\documentclass[10pt,twocolumn,letterpaper]{article}

\usepackage{cvpr}              

\usepackage[table]{xcolor}
\definecolor{mydarkblue}{rgb}{0,0.08,0.45}
\usepackage[
            breaklinks,
            colorlinks,
            linkcolor=mydarkblue,
            citecolor=mydarkblue,
            filecolor=mydarkblue,
            urlcolor=mydarkblue,]{hyperref}

\usepackage{graphicx}
\usepackage{amsmath}
\usepackage{amssymb}
\usepackage{booktabs}
\usepackage{mathtools}
\usepackage{xcolor}
\usepackage{color, colortbl}
\usepackage{float}
\usepackage{multirow}
\usepackage{threeparttable}
\usepackage{fancyvrb}
\usepackage{caption}
\usepackage{soul}
\usepackage{diagbox}

\usepackage{microtype}
\definecolor{Gray}{gray}{0.9}

\usepackage[capitalize]{cleveref}
\crefname{section}{Sec.}{Secs.}
\Crefname{section}{Section}{Sections}
\Crefname{table}{Table}{Tables}
\crefname{table}{Tab.}{Tabs.}

\def\cvprPaperID{1319} 
\def\confName{CVPR}
\def\confYear{2022}

\begin{document}

\title{Audio-Adaptive Activity Recognition Across Video Domains}

\author{Yunhua Zhang$^{1}$~~~~Hazel Doughty$^{1}$~~~~Ling Shao$^{2}$\thanks{Currently at Terminus Group, China.}~~~~Cees G. M. Snoek$^{1}$\\[1mm]
\normalsize $^{1}$University of Amsterdam~~~$^{2}$Inception Institute of Artificial Intelligence \\
}

\maketitle


\begin{abstract}
\vspace{-0.8em}
   This paper strives for activity recognition under domain shift, for example caused by change of scenery or camera viewpoint. The leading approaches reduce the shift in activity appearance by adversarial training and self-supervised learning. 
   Different from these vision-focused works we leverage activity sounds for domain adaptation as they have less variance across domains and can reliably indicate which activities are not happening. 
   We propose an audio-adaptive encoder and associated learning methods that discriminatively adjust the visual feature representation as well as addressing shifts in the semantic distribution. To further eliminate domain-specific features and include domain-invariant activity sounds for recognition, an audio-infused recognizer is proposed, which effectively models the cross-modal interaction across domains. 
   We also introduce the new task of actor shift, with a corresponding audio-visual dataset, to challenge our method with situations where the activity appearance changes dramatically. Experiments on this dataset, EPIC-Kitchens and CharadesEgo show the effectiveness of our approach.
   Project page: \url{https://xiaobai1217.github.io/DomainAdaptation}. 
\end{abstract}
\vspace{-1em}

\section{Introduction}
\begin{figure}[t!]
\centering 
\includegraphics[width=0.95\linewidth,height=0.9\linewidth]{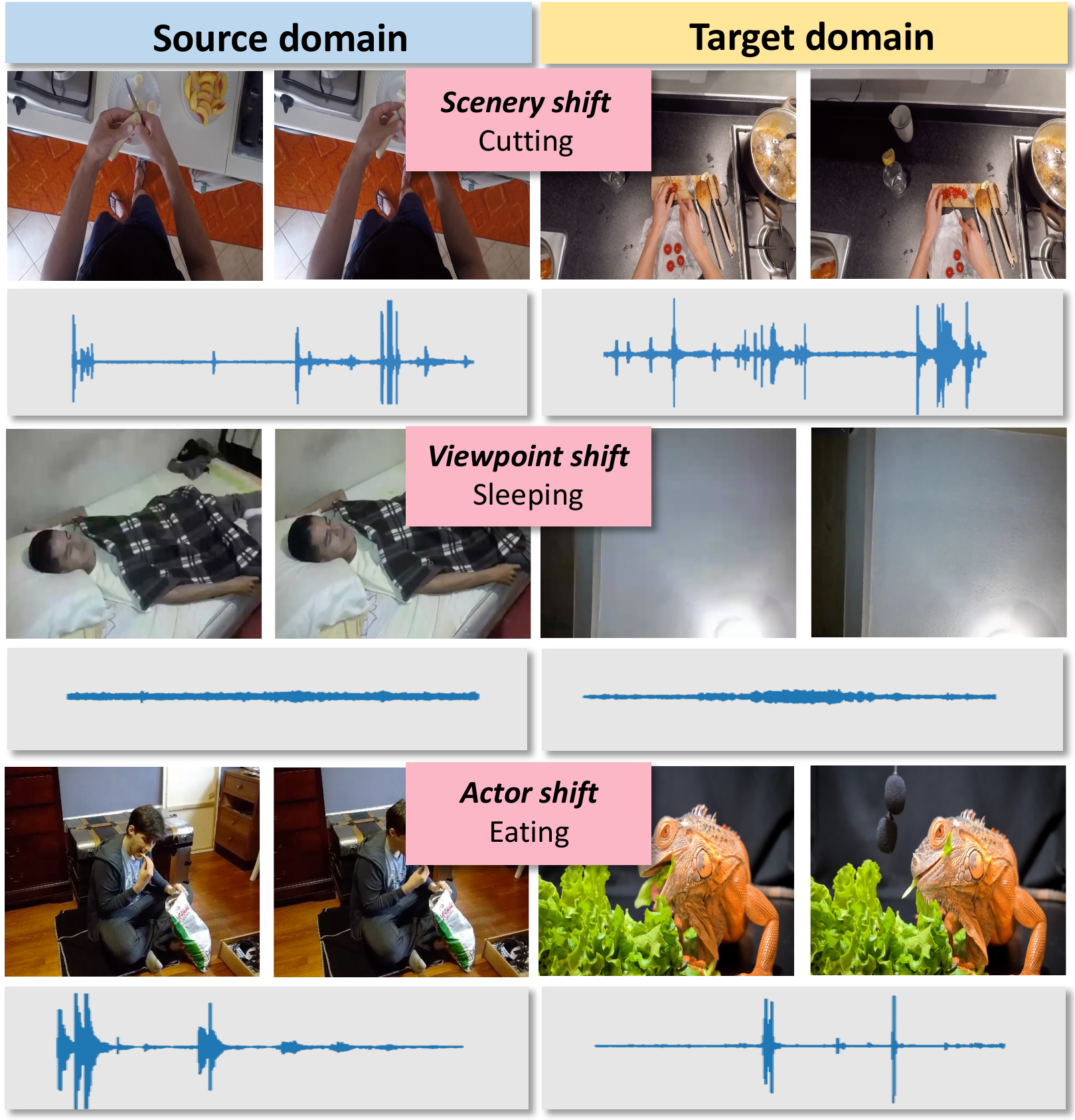}
\vspace{-0.5em}
\caption{
We recognize activities under domain shifts, caused by change of scenery, camera viewpoint or actor, with the aid of sound. }
\label{fig:1st_figure}
\vspace{-1.2em}
\end{figure}

\vspace{-0.3em}
The goal of this paper is to recognize activities such as \emph{eating}, \emph{sleeping} or \emph{cutting} under domain shift caused by change of scenery, camera viewpoint or actor, as shown in Figure~\ref{fig:1st_figure}. Existing solutions align distribution-shifted domains inside a single visual video network by adversarial training~\cite{munro2020multi,pan2020adversarial,jamal2018deep,chen2019temporal} and self-supervised learning~\cite{song2021spatio,choi2020shuffle,iccv2021videoadaptation}. Although successful, projecting the visual features from different source and target domains into a shared space can make the ability of the model to distinguish between classes in the target domain suffer. We observe that activity sounds can
act as natural domain-invariant cues, as they carry rich activity information while exhibiting less variance across domains. We thus propose a video model which adapts to video distribution shifts with the aid of sound.

Many have considered sound in addition to visual analysis for activity recognition within a single domain~\cite{lee2021crossattentional,wu2019dual,gao2020listen,korbar2019scsampler,tian2018audio,tian2020unified,zhang2021repetitive,wu2021exploring,rai2021home,nagrani2021attention}.
For instance, both Gao \etal \cite{gao2020listen} and Korbar \etal \cite{korbar2019scsampler} reduce the computational cost by previewing the audio track, while Lee \etal \cite{lee2021crossattentional} show that combining visual features with audio can better localize actions. However, the cross-modal correspondences become harder to discover when shifting domains, causing existing cross-modal fusion schemes to degrade in performance. Yang \etal \cite{yang2021epic} and Planamente \etal~\cite{planamente2021cross} propose to directly fuse visual and audio features or predictions for cross-domain activity classification. 
However, the effectiveness of these methods is reduced when not all activities make a characteristic sound. 
Different from previous works, we introduce audio-adaptive learning methods and a cross-modal interaction that utilizes the reliable domain-invariant cues within sound to help the video model adapt to the distribution shift. 

We make three contributions in this paper. First, we propose an audio-adaptive encoder which exploits the rich information from sound to adjust the visual feature representation causing the model to learn more discriminative features in the target domain. This is done by preventing the model from over-fitting to domain-specific visual content, while simultaneously dealing with imbalanced semantic distributions between domains. Second, we introduce an audio-infused recognizer, which eliminates domain-specific features further and allows effective cross-modal interaction across domains by considering domain-invariant activity information within sound. As a third contribution, we introduce the new task of \textit{actor shift}, and a corresponding audio-visual video dataset \emph{ActorShift}, to challenge our approach when the change in actors results in large variation in activity appearance. Experiments on EPIC-Kitchens~\cite{damen2018scaling}, CharadesEgo~\cite{sigurdsson2018actor} and \textit{ActorShift}, demonstrate the advantage of our approach under various video distribution shifts for both audible and silent activities. 

\section{Related Work}
\vspace{-0.3em}
\noindent \textbf{Sound for activity recognition.} Many works have utilized sound for within-domain activity recognition in videos, \eg,~\cite{gao2020listen,kazakos2019epic,korbar2019scsampler,tian2018audio,tian2020unified,lee2021crossattentional}. Since there is a natural correlation between the visual and auditive elements of a video, Korbar \etal~\cite{korbar2018cooperative} and Asano \etal~\cite{asano2020labelling} learn audio-visual models in a self-supervised manner. As processing audio signals is much faster than video frames, both Gao~\etal~\cite{gao2020listen} and Korbar~\etal\cite{korbar2019scsampler} reduce computation by previewing the audio track for video analysis. Cross-modal attention is widely used in activity localization~\cite{lee2021crossattentional,wu2019dual,tian2018audio} and audiovisual video parsing~\cite{wu2021exploring,tian2020unified} to guide the visual model to focus on the audible regions. Zhang \etal~\cite{zhang2021repetitive} conduct repetitive activity counting by using audio signals to decide the sampling rate and predict the reliability of the visual features. As opposed to most works which rely on sound for within-domain activity recognition, we consider its domain-invariant nature for activity recognition across different domains.

\noindent \textbf{Video domain adaptation by vision. }
The field of vision-focused domain adaptation is extensive (see recent surveys 
~\cite{wang2018deep,zhuang2020comprehensive}). Here, we focus on video domain adaptation for activity recognition. State-of-the-art visual-only solutions learn to reduce the shift in activity appearance by adversarial training~\cite{jamal2018deep,chen2019temporal,pan2020adversarial,munro2020multi,choi2020shuffle,choi2020unsupervised,chen2020action} and self-supervised learning techniques~\cite{munro2020multi,choi2020shuffle,song2021spatio,iccv2021videoadaptation}. While Jamal \etal~\cite{jamal2018deep} and Munro and Damen~\cite{munro2020multi} directly penalize domain specific features with an adversarial loss at every time stamp, Chen \etal~\cite{chen2019temporal}, Choi \etal~\cite{choi2020shuffle} and Pan \etal~\cite{pan2020adversarial} attend to temporal segments that contain important cues. Self-supervised learning objectives are also incorporated in \cite{munro2020multi} and \cite{choi2020shuffle} to better align the features across domains by utilizing the correspondences between RGB and optical flow or the temporal order of video clips. Song~\etal~\cite{song2021spatio} and Kim~\etal~\cite{iccv2021videoadaptation} obtain remarkable performance by contrastive learning for self-supervised learning to align the feature distributions between video domains. Instead of relying on the vision modality only, which may present large activity appearance variance, we consider the domain-invariant information within sound to help the model adapt to the visual distribution shift. 

\noindent \textbf{Video domain adaptation by vision and audio. }
As audio signals contain valuable domain-invariant cues, some recent works recognize activities across domains with the aid of sound. Yang \etal~\cite{yang2021epic} directly fuse the features from visual and audio modalities before classification. 
However, this can lead to the visual features dominating the classification since many activities are silent and the audio features are less discriminative. 
As a result, the complementary information from sound may not be considered. 
Planamente \etal~\cite{planamente2021cross} instead align the two modalities with an audio-visual loss. 
Nonetheless, the audio predictions for silent activities remain unreliable and limit their performance improvements. 
Instead, 
we propose audio-adaptive learning that exploits the supervisory signals from sound to adjust to the distribution shift and handle both audible and silent activities. 

Additionally, existing datasets \eg,~\cite{damen2018scaling,sigurdsson2018actor,Kinetics,soomro2012ucf101} focus on human actors, 
meaning activities are inherently close in appearance and share 
commonalities with hand-object interactions. 
Inspired by the A2D dataset by Xu \etal~\cite{xu2015can}, which contains multiple actor classes for activity recognition, we introduce the challenging domain adaptation setting of \emph{actor shift}, in which the shift between humans and animals performing the action results in large appearance and motion differences across domains, further facilitating video domain adaptation by the use of vision and audio.

\begin{figure*}[t!]
\centering
\includegraphics[width=0.9\linewidth,height=0.35\linewidth]{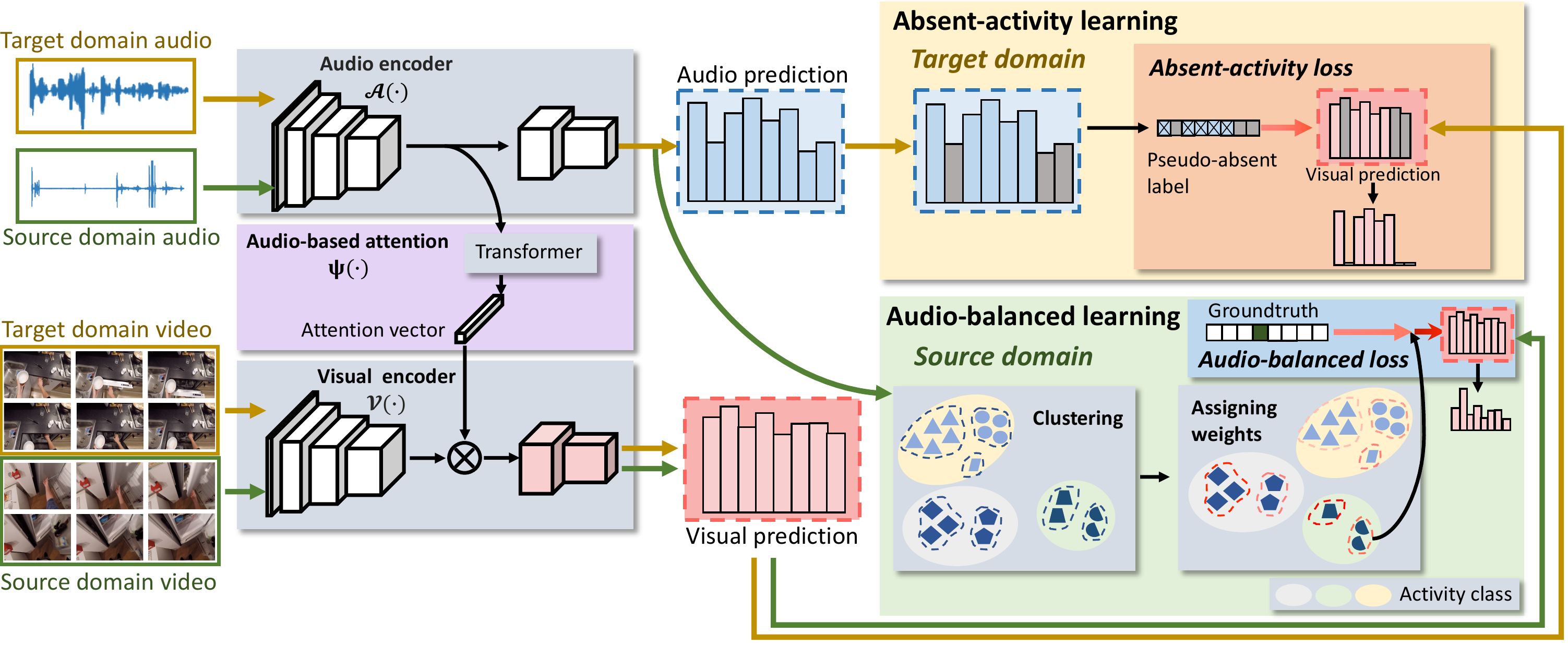}
\vspace{-0.5em}
\caption{\textbf{Audio-adaptive encoder for activity recognition under domain shift.} With a pre-trained audio encoder, we train the visual encoder and audio-based attention module, which guides the visual encoder to focus on the activity relevant features. We do this with two audio-adaptive learning methods: absent-activity learning and audio-balanced learning. The absent activity learning operates in the target domain and uses the audio predictions to indicate which activities cannot be heard in the video. The visual predictions are then encouraged to have low probabilities for these `pseudo-absent' activities. The audio-balanced learning uses audio in the source domain to cluster samples in each activity class into clusters according to the sounds of the object/environment interacted with. In the audio-balanced loss the rare activities and interactions are weighted higher to handle the semantic shift between domains. 
}
\vspace{-1em}
\label{fig:framework}
\end{figure*}

\section{Approach}
\vspace{-0.3em}
\label{sec:approach}

For activity recognition under domain shift, we consider unsupervised domain adaptation where we have: 
a set of labeled source videos 
$\mathcal{S}{=}\{(X_1^\mathcal{S}, y_1^{\mathcal{S}}),\dots,(X_{N}^\mathcal{S}, y_{N}^{\mathcal{S}})\}$ and 
a set of unlabeled target videos $\mathcal{T}{=}\{X_1^{\mathcal{T}},\dots,X_{M}^{\mathcal{T}}\}$. 
In each domain, $X$ and $y$ indicate a video sample and the corresponding activity class label, while $N$ and $M$ are the number of samples in the source and target domain. 
Using all available training data from the source and the target domains, the task is to train an activity recognition model, which performs well on (unseen) videos from the target domain.

We train our audio-adaptive model in two stages using videos from source and target domains with accompanying audio. In the first stage we train our audio-adaptive encoder (Section~\ref{sec:encoder}) that uses audio to adapt a visual encoder to be more robust to distribution shifts. In the second stage we train our audio-infused recognizer (Section~\ref{sec:transformer}) using pseudo-labels from the audio-adaptive encoder for the target domain and the ground-truth labels for the source domain. The audio-infused recognizer maps the source and target domains into a common space and fuses audio and visual features to produce an activity prediction for either domain.

\subsection{Stage 1: Audio-Adaptive Encoder}
\vspace{-0.3em}
\label{sec:encoder}

Our audio-adaptive encoder $\mathcal{E}(\cdot)$, detailed in Figure~\ref{fig:framework}, consists of a visual encoder $\mathcal{V}(\cdot)$, an audio encoder $\mathcal{A}(\cdot)$ and an audio-based attention module $\psi(\cdot)$. 
Since the sounds of activities have less variance across domains, $\mathcal{E}(\cdot)$ aims to extract visual features that are invariant but discriminative under domain shift with the aid of $\mathcal{A}(\cdot)$ pre-trained for audio-based activity recognition. 
To this end, we train $\mathcal{V}(\cdot)$ and $\psi(\cdot)$ with two audio-adaptive learning methods: 
absent-activity learning for unlabeled target data and audio-balanced learning for labeled source data. 
The former aims to remove irrelevant parts of the visual features while the latter helps to handle the differing label distribution between domains. 
Once trained, for each video, 
we can extract an audio feature vector from $\mathcal{A}(\cdot)$ and a series of visual features from $\mathcal{V}(\cdot)$ with which to train our audio-infused recognizer (Section~\ref{sec:transformer}) for activity classification.

\noindent\textbf{Audio-based attention.} We use an audio-based attention module $\psi(\cdot)$ to adapt the 
visual encoder to focus on activity-relevant features. For example, the visual model may predict the activity \emph{washing} because of the presence of a sink. However, without the sound of water the attention module suppresses the channels encoding the sink thus increasing the prediction 
of the correct class. The attention module is based on the 
transformer encoder\cite{vaswani2017attention,devlin2018bert,vit}. It takes the audio features as input and outputs the channel attention feature vector, which is 
multiplied with the visual features.

\noindent \textbf{Absent-activity learning.}
The absent-activity learning uses audio in the target domain to train the attention module and visual encoder. Naively, we could treat the class with the highest probability from the visual encoder as the pseudo label. 
However, 
doing so can create biased pseudo-labels as irrelevant objects often appear in a scene. Instead, we use the audio predictions to guide the visual pseudo-labels. 
While we may not be confident which activity is happening in a video, particularly for silent videos, 
we can often be confident that certain activities with distinctive sounds are \emph{not} occurring in a video. We call these ``absent activities". 
To learn from these absent activities, we generate pseudo-absent labels for the unlabeled target domain videos, which indicate the activities with the lowest probabilities from the audio encoder. The visual encoder is then encouraged to predict these unlikely classes with low probability. 

Specifically, for an unlabeled video $X^{\mathcal{T}}$ in the target domain, we obtain the audio-based activity 
probability distribution $\mathbf{p}_{a}^{\mathcal{T}} \in \mathbb{R}^K$ ($K$ is the number of classes) from the audio encoder $\mathcal{A}(\cdot)$ trained on labeled source data. From this we obtain the set of absent activities $\mathcal{Q}$ by taking the lowest $r$ predictions in $\mathbf{p}_{a}^\mathcal{T}$, \ie, the classes with the lowest probabilities from the audio encoder. We also extend this to multi-label classification by instead assuming the $(1-\alpha_k)\gamma$ percent videos with the lowest probabilities do not contain class $k$, where $\gamma \in (0,1]$ and $\alpha_k$ is the percentage of videos containing each activity class in the labeled source domain.

Our loss for absent-activity learning is formulated as: 
\begin{equation}
    l_{\text{A}}(\textbf{p}_{v}^{\mathcal{T}}, \mathcal{Q}) = -\sum_{q \in \mathcal{Q}}\log(1-p^{\mathcal{T}}_{{v,q}}), 
\end{equation}
where 
$p^{\mathcal{T}}_{{v,q}}$ is the probability output for the $q$th class for the video $X^\mathcal{T}$. 
With this loss, the visual encoder is able to ignore confounding visual features and generate less-noisy pseudo-labels for the target domain. This allows our model to better capture high-level semantic information between domains based on both appearance and motion cues. 

\noindent \textbf{Audio-balanced learning.} Besides a change in visual appearance, domain shift can also be caused by a change in 
label distributions~\cite{munro2020multi} and frequencies of objects/environments. 
For example, the \emph{open} activity may commonly occur on a `cupboard' in the source domain but be more common with a `can' in the target. These two cases result in different audio-visual activity appearances. 
We address such challenges with our audio-balanced learning, which not only handles imbalance in activity classes, but also imbalance in terms of 
the objects or the environment being interacted with.

To this end, we first use $k$-means to group the video samples inside each activity class by their audio feature $\mathbf{f}^\mathcal{S}_{a}$ 
with the assumption that each group represents a different type of object or environment.
We use audio features for clustering as they can indicate the material of the interacted objects or the environment the action is performed in, while being invariant to appearance changes.
The number of interaction clusters per activity class is determined by the Elbow method~\cite{thorndike1953belongs}, which favours a small number while obtaining a low ratio of dispersion both between and within clusters.

We based our \textit{audio-balanced} loss on the class-balanced loss 
by Cui \etal~\cite{cui2019class}. When using the original class-balanced loss on a source domain video $X^\mathcal{S}$ with visual probabilities $\mathbf{p}_v^\mathcal{S}$ we can balance over our activity classes:
\vspace{-0.5em}
\begin{equation}
l_{\text{CB}}(\mathbf{p}_{v}^{\mathcal{S}}, y^{\mathcal{S}})
= \frac{1 - \beta}{1 - \beta^{n_{y}}}\ \mathcal{L}(\mathbf{p}_{v}^{\mathcal{S}}, y^{\mathcal{S}}),
\label{eq:CB}
\end{equation}
\noindent where $\mathcal{L}$ is a classification loss, \eg, softmax cross-entropy loss and $n_{y}$ is the number of training samples of ground-truth activity class $y$. $\beta \in [0,1)$ is a hyper-parameter which controls the weighting factor $\frac{1-\beta}{1-\beta^{n_{y}}}$. As $\beta\rightarrow1$, this weighting factor becomes inversely proportional to the effective number of samples inside each class so that tail classes in the source domain are weighted higher in training.

With our \textit{audio-balanced loss} we include an additional weighting factor so the long tail of object interactions are also accounted for with our interaction clusters: 
\begin{equation}
    l_{\text{B}}(\mathbf{p}_{v}^{\mathcal{S}},y^{\mathcal{S}}) = \frac{1-\beta}{1-\beta^{n_{y,j}}} l_{\text{CB}}(\mathbf{p}_{v}^{\mathcal{S}},y^{\mathcal{S}}).
\end{equation}
$n_{y,j}$ is the number of samples for the $j$th interaction cluster that video $X^{\mathcal{S}}$ is assigned within ground-truth activity $y^{\mathcal{S}}$. By this loss, both rare activities and rare interactions from frequent activities are given a high weight during training. This means the classifier can generalize well to the target domain where the distribution of activities and interactions may not be the same.

\noindent \textbf{Audio-adaptive encoder loss.} The absent-activity loss and the audio-balanced loss are combined to obtain the overall loss for training the visual encoder $\mathcal{V}(\cdot)$ and audio-based attention $\mathcal{\psi}(\cdot)$ inside the audio-adaptive encoder $\mathcal{E}(\cdot)$: 
\begin{equation}
\label{eq:unsupervised_loss}
    l_{\mathcal{E}} = \sum_{(X_i)\in\mathcal{T}}l_{\text{A}}(\textbf{p}_{i,v}^{\mathcal{T}}, \mathcal{Q}_{i}) + \smashoperator{\sum_{(X_j,y_j)\in\mathcal{T}}}l_{\text{B}}(\textbf{p}_{j,v}^{\mathcal{S}}, y_{j}^{\mathcal{S}}).
\end{equation}
%


\begin{figure}[t!]
\centering
\includegraphics[width=0.9\linewidth,height=0.5\linewidth]{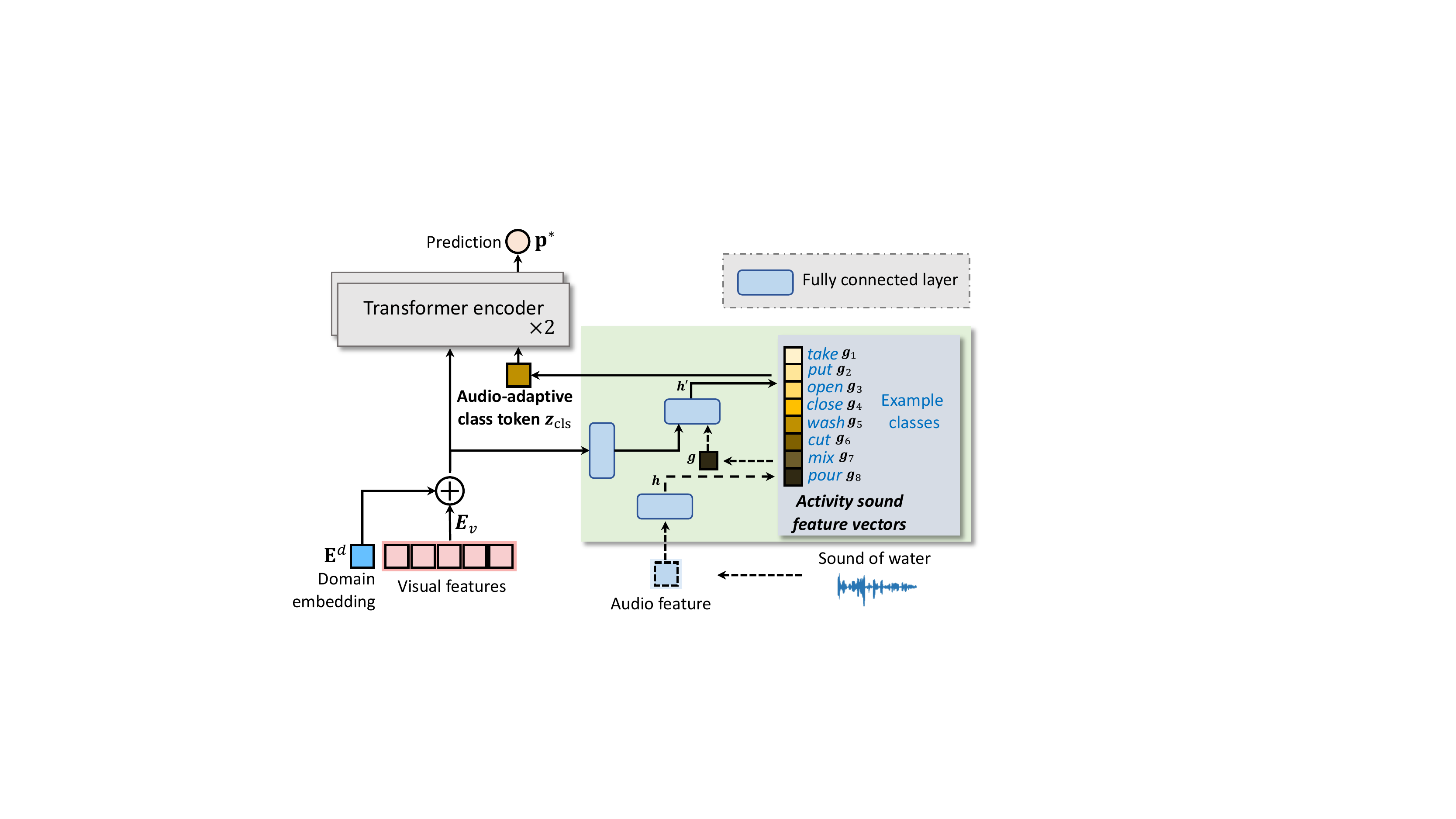}
\vspace{-0.5em}
\caption{\textbf{Audio-infused recognizer.} We add domain embedding $E_d$ to encourage a common visual representation across domains. 
Then, an audio-adaptive class token is obtained from a series of activity sound feature vectors, considering both audio and visual features. 
It is sent into the transformer together with the visual features. 
By the transformer's self attention, 
this token aggregates information from visual features with the domain-invariant audio activity cues for activity classification. 
}
\vspace{-0.5em}
\label{fig:transformer}
\end{figure}

\subsection{Stage 2: Audio-Infused Recognizer}
\label{sec:transformer}
\vspace{-0.3em}
While audio can help focus on the activity-relevant visual features, there is still a large difference between the appearance of activities in different domains. To further eliminate domain-specific visual features and fuse the activity cues from the audio and visual modalities we propose the audio-infused recognizer $\mathcal{R}(\cdot)$, visualized in Figure~\ref{fig:transformer}.

\begin{table*}[th!]
\renewcommand{\arraystretch}{1.1}
\centering
\resizebox{0.8\linewidth}{!}{
\begin{tabular}{lllrlrr}
\toprule
&  & \multicolumn{2}{c}{\textbf{Source Domain Setting}} &\multicolumn{3}{c}{\textbf{Target Domain Setting}}\\
 \cmidrule(lr){3-4} \cmidrule(lr){5-7}
\textbf{Shift} & \textbf{Video Dataset} & \textit{Source Domain} & \textit{Train} & \textit{Target Domain} & \textit{Train} & \textit{Test}\\
\midrule
\textbf{Scenery} & EPIC-Kitchens-55~\cite{damen2018scaling} & Kitchens & 7,935 & Kitchens & 7,935 & 2,114 \\
\textbf{Viewpoint} & CharadesEgo~\cite{sigurdsson2018actor} & Third-person view & 3,083 & Ego-centric view & 3,083 & 825 \\
\textbf{Actor} & ActorShift (\textit{ours}) & Human actors & 1,305 & Animal actors & 35 & 165 \\
\bottomrule
\end{tabular}
}
\vspace{-0.8em}
\caption{\textbf{Domain adaptation benchmarks for activity recognition under scenery, viewpoint and actor shift} with the datasets used and number of videos per source and target split. Scenery and viewpoint shift are present in existing datasets.
We propose the actor shift setting and dataset to tackle the challenge of a severe change in activity appearance. The dataset is available on the project website. 
}
\vspace{-1em}
\label{tab:datasets}
\end{table*}

\noindent\textbf{Transformer with domain embedding.} We adopt a transformer encoder since its core mechanism, self-attention, can efficiently encode multi-modal representations~\cite{gabeur2020multi,zhu2020actbert,sun2019videobert}. 
For a vanilla version, we take the input sequence: 
\begin{equation}
\label{eq:zm}
\mathbf{z}^{m} {=} [\mathbf{z}_{\text{cls}}^{m}; \mathbf{f}_{v,1}\mathbf{E}_v;\cdots;\mathbf{f}_{v,n}\mathbf{E}_v;\mathbf{f}_{a,1}\mathbf{E}_a,;\cdots;\mathbf{f}_{a,n}\mathbf{E}_a],
\end{equation}
where $\mathbf{z}_{\text{cls}}^{m}$ is the learnable class token defined as in \cite{vit}, and $\{\mathbf{f}_{v,1}, \cdots, \mathbf{f}_{v,n}| \mathbf{f}_{v,\cdot} \in \mathbb{R}^{C_v}\}$  and $\{\mathbf{f}_{a,1}, \cdots, \mathbf{f}_{a,n} | \mathbf{f}_{a,\cdot} \in \mathbb{R}^{C_a}\}$ are the visual and audio features of $n$ clips from video $X$. 
$\mathbf{E}_v \in \mathbb{R}^{C_v \times D}$ and $\mathbf{E}_a \in \mathbb{R}^{C_a \times D}$ are linear projections to map the visual and audio features to $D$ dimensions. 
To map source and target domains into a common space, we first learn a domain embedding $\mathbf{E}^d \in \mathbb{R}^D$ ($d \in \{\mathcal{S}, \mathcal{T}\}$), which contains both positive and negative values and is added to suppress domain-specific visual features. Then, the input sequence for the transformer becomes: 
\begin{equation}
\label{eq:z'}
    \mathbf{z}' {=} [\mathbf{z}_{\text{cls}}^{m}; \mathbf{f}_{v,1}\mathbf{E}_v+\mathbf{E}^d;{\cdots};\mathbf{f}_{v,n}\mathbf{E}_v+\mathbf{E}^d;\mathbf{f}_{a,1}\mathbf{E}_a,;{\cdots};\mathbf{f}_{a,n}\mathbf{E}_a]. 
\end{equation}

\vspace{-0.2em}
\noindent\textbf{Audio-adaptive class token. }
Ideally, the transformer's self attention will aggregate audio and visual features with the class token to predict the correct activity. 
However, the cross-modal correspondences are difficult to find under distribution shift, meaning the prediction may rely on the more discriminative, but less domain-invariant, visual features. 
To address this, we propose to generate an audio-adaptive class token, which is initialized from the audio activity class prediction and gradually aggregates the visual features while keeping its own audio-based activity information through the transformer. 
As shown in Figure~\ref{fig:transformer}, the audio-adaptive class token is obtained from a series of activity sound vectors $\{\mathbf{g}_k \in \mathbb{R}^{D}\}^K_{k=1}$, with each representing an activity class. They capture global context information and serve as the representation bottleneck to provide regularization for model learning~\cite{belghazi2018mutual,rame2021dice}. 
For selection, the feature vector from the audio adaptive encoder $\mathcal{A}(X)$ is first processed by a fully connected layer to give the activity probabilities $\mathbf{h} \in \mathbb{R}^K$. 
Then, an initial vector is obtained by $\mathbf{g} {=} \sum_{k=1}^Kh_k * \mathbf{g}_k$. 
We include visual features to help silent activities select the representative vector. 
To avoid the visual features dominating, we project them to a lower dimension with a fully connected layer before concatenating them with the initial vector $\mathbf{g}$. 
The concatenated vector is given to another fully connected layer which outputs the probabilities $\mathbf{h}'$ for each type of activity sound. Finally, we obtain the audio representation $\mathbf{z}_{\text{cls}} {=} \sum_{k=1}^Kh_k' * \mathbf{g}_k$, which serves as the class token. 
Consequently, the input sequence for the transformer becomes:
\vspace{-0.3em}
\begin{equation}
\label{eq:z}
\mathbf{z} {=} [\mathbf{z}_{\text{cls}}; \mathbf{f}_{v,1}\mathbf{E}_v+\mathbf{E}^d,;\cdots;\mathbf{f}_{v,n}\mathbf{E}_v+\mathbf{E}^d],
\end{equation}
where $\mathbf{z}_{\text{cls}}$ is the audio-adaptive class token. 
The class token output state is further sent to a fully connected layer to get the final prediction $\mathbf{p}^*$.
For audible activities, the activity sound vector can be accurately selected and kept discriminative for audiovisual interaction. 
For silent activities, the vector is obtained from environmental sound, which indicates the presence of multiple possible activities. The vector becomes more discriminative as 
the transformer progressively enhances it through the 
visual features.

\noindent \textbf{Audio-infused recognizer loss.}
We train the audio-infused recognizer on both source and target videos with the loss:
\begin{equation}
\label{eq:loss_transformer}
\begin{split}
        l_{\mathcal{R}} =  \smashoperator{\sum_{(X_i, y_i)\in\{\mathcal{S},\mathcal{T}\}}}  \mathcal{L}(\mathbf{p}_i^*, y_i) 
         + \eta \Big( \mathcal{L}(\mathbf{h}_i, y_i) + \mathcal{L}(\mathbf{h}_i', y_i) \Big ),  
\end{split}
\end{equation}
where hyperparameter $\eta$ balances the loss terms and $y_i$ is the groundtruth or, in the case of the unlabeled video, the hard pseudo-label. 
$\mathbf{p}_i^*$ is the final classification prediction, and $\mathbf{h}_i$ and $\mathbf{h}_i'$ are the probabilities for the activity sound vectors outputted by the first and second fully connected layers. 
The first term $\mathcal{L}(\mathbf{p}_i^*, y_i)$ optimizes the transformer to predict the correct activity class, while the second term $\mathcal{L}(\mathbf{h}_i, y_i) + \mathcal{L}(\mathbf{h}_i', y_i)$ optimizes the 
activity sound vectors. 
We are now ready to validate the effectiveness of our approach on three domain adaptation benchmarks as highlighted in Figure~\ref{fig:1st_figure}, summarized in Table~\ref{tab:datasets} and detailed next.

\section{Domain Adaptation Benchmarks}
\vspace{-0.3em}
\noindent \textbf{Scenery shift.}
We study scenery shift in the \textit{\textbf{EPIC-Kitchens-55}}~\cite{damen2018scaling} dataset, which contains first-person videos of fine-grained kitchen activities. The domain adaptation benchmark proposed by Munro and Damen 
\cite{munro2020multi} uses three domain partitions (D1, D2 and D3), 
where each domain is a different person in a different kitchen. The task is to adapt between each pair of domains. This benchmark focuses on eight activity classes (verbs), which occur in combination with different objects, with a severe class imbalance. The kitchens have different appearances and contain different utensils. 

\noindent \textbf{Viewpoint shift.} We consider viewpoint shift in the \textit{\textbf{CharadesEgo}} dataset by Sigurdsson \etal~\cite{sigurdsson2018actor}. It contains paired videos of the same activities, recorded from first and third-person perspective. It has 3,083 and 825 videos per viewpoint for training and testing, spanning 157 activity classes. Following ~\cite{choi2020unsupervised}, we treat the third-person videos as the source domain and the first-person videos as the target domain. The changing views make the activities appear visually different, resulting in a large domain gap. 

\noindent \textbf{Actor shift.} While both EPIC-Kitchens and CharadesEgo contain considerable domain shifts, there are still some inherent similarities between the domains in these datasets. Since all the actors are humans, latent signals describing the way hands and objects interact are shared between domains. Therefore, we introduce an even more challenging domain shift setting to further facilitate video domain adaption research and demonstrate the potential of our method. We introduce \emph{ActorShift}, where the domain shift comes from the change in actor species: we use humans in the source domain and animals in the target domain. This causes large variances in the appearance and motion of  activities. 

For the corresponding dataset we select 1,305 videos of 7 human activity classes from Kinetics-700~\cite{Kinetics} as the source domain: \emph{sleeping}, \emph{watching tv}, \emph{eating}, \emph{drinking}, \emph{swimming}, \emph{running} and \emph{opening a door}. For the target domain we collect 200 videos from YouTube of animals performing the same activities. 
We divide them into 35 videos for training (5 per class) and 165 for evaluation. 
The target domain data is scarce, meaning there is the additional challenge of adapting to the target domain with few unlabeled examples. 

\noindent \textbf{Evaluation criteria.} Following standard practice~\cite{munro2020multi,sigurdsson2018actor}, we report top-1 accuracy on EPIC-Kitchens and ActorShift for single-label classification, and mAP (mean average precision) on CharadesEgo for multi-label classification.

\section{Results}
\vspace{-0.3em}
We first describe the implementation details before ablating the components of our method and comparing to prior works for each type of domain shift.

\noindent \textbf{Implementation details.} For our visual encoder $\mathcal{V}(\cdot)$ we use SlowFast~\cite{feichtenhofer2019slowfast}, unless stated otherwise. 
For the audio encoder $\mathcal{A}(\cdot)$ we use ResNet-18~\cite{resnet}. The audio-based attention module $\psi(\cdot)$ consists of eight transformer encoder layers~\cite{vit} with a final fully connected layer to obtain the attention vector for the visual encoder. The inputs are intermediate audio features from $\mathcal{A}(\cdot)$ (conv3) along with a learnable class token defined as in~\cite{vit} (note this is different from our audio-adaptive class token used in $\mathcal{R}(\cdot)$). 
The output state of the class token passes through the fully connected layer to obtain the attention vector for $\mathcal{V}(\cdot)$. 
We set the parameters of our absent activity loss to $r{=}3$, $\gamma{=}0.05$ and $\beta{=}0.999$. Our audio-infused recognizer $\mathcal{R}(\cdot)$ consists of two transformer encoder layers~\cite{vit} and three fully connected layers for generating the class token. The sequence dimension $D$ is 512 and each layer has 8 self-attention heads. More details are in the supplementary. 

\subsection{Ablation Study}
\vspace{-0.3em}
For ablations we use RGB and audio modalities on both EPIC-Kitchens and CharadesEgo. During training, all labeled source videos are used. With EPIC-Kitchens all target videos are unlabelled, while for CharadesEgo we use half 
labelled and half unlabelled for semi-supervised domain adaptation as in~\cite{choi2020unsupervised}. Since EPIC-Kitchens contains multiple adaptation settings, we report the average. Ablations on component internals are provided in the supplementary.

\begin{table}[t!]
\renewcommand{\arraystretch}{1.1}
\centering
\resizebox{1\linewidth}{!}{
\begin{threeparttable}
\begin{tabular}{lcc}
\toprule
  & \multicolumn{1}{c}{\textbf{EPIC-Kitchens}} & \multicolumn{1}{c}{\textbf{CharadesEgo}} \\
\cmidrule(lr){2-2} \cmidrule(lr){3-3}
 \textbf{Model} & Top-1 (\%) $\uparrow$ & mAP (\%) $\uparrow$  \\
\midrule
\rowcolor{Gray}
\textbf{Stage 1: Audio-adaptive encoder $\mathcal{E}(\cdot)$}&  &  \\
Visual encoder $\mathcal{V}(\cdot)$ & 48.0  & 23.1  \\ 
+ Audio-based attention $\psi(\cdot)$ & 51.2 & 23.5 \\
+ Absent-activity learning & 53.7 & 24.4 \\
+ Audio-balanced learning & 55.7 & 25.0\\
\rowcolor{Gray}
\textbf{Stage 2: Audio-infused recognizer $\mathcal{R}(\cdot)$} &  &  \\
+ Vanilla multi-modal transformer $\mathbf{z}^{m}$  & 56.1 & 25.0 \\ 
+ Domain embedding $\mathbf{z}'$ & 57.2 & 25.4 \\
+ Audio-adaptive class token $\mathbf{z}$ & 59.2 & 26.3 \\
\bottomrule
\end{tabular}
\end{threeparttable}
}
\vspace{-1em}
\caption{\textbf{Model components ablation.} All components in 
the audio-adaptive encoder and the audio-infused recognizer contribute to performance improvement under distribution shift. For both EPIC-Kitchens and CharadesEgo the improvements over a vanilla SlowFast visual encoder are considerable. }
\vspace{-0.8em}
\label{tab:new_module_ablation}
\end{table}

\noindent \textbf{Stage 1: Audio-adaptive encoder.} We report results in Table~\ref{tab:new_module_ablation}. We first consider the audio-adaptive encoder alone. Initially, we train only the visual encoder with a standard softmax cross-entropy loss on the source domain. Simply generating channel attention for the visual features with our audio-based attention module already improves performance by 3.2\% top-1 accuracy on EPIC-Kitchens and 0.4\% mAP on CharadesEgo. Since audio contains useful activity information, this attention helps the visual encoder focus on relevant features. Adding the absent-activity learning results in 2.5\% and 0.9\% improvements, demonstrating that the pseudo-absent labels increase the discriminative ability of the model in the target domain. We observe that adopting the audio-balanced learning and replacing the softmax cross-entropy with our audio-balanced loss delivers a further 2.0\% and 0.6\% increase. This highlights the importance of addressing the label distribution shift in domain adaption. 

\noindent \textbf{Stage 2: Audio-infused recognizer.} 
For the audio-infused recognizer, we first consider a vanilla transformer. It takes as input $\mathbf{z}^{m}$ (Eq.~\ref{eq:zm}), \ie the audio and visual features from the audio-adaptive encoder, mapped by $\mathbf{E}_v$ and $\mathbf{E_a}$ into a common space, alongside a learnable class token. 
This only gives a marginal improvement in results. Adding the domain embedding $\mathbf{E}^d$ to reduce domain-specific visual features in $\mathbf{z}'$ (Eq.~\ref{eq:z'}) gives a benefit of 1.1\% on EPIC-Kitchens and 0.4\% on CharadesEgo. This is because the cross-modal correspondences become easier to discover. When we replace the plain audio features and single learnable class token with our audio-adaptive class token to get $\mathbf{z}$
(Eq.~\ref{eq:z}), we observe further improvements of 2.0\% and 0.9\%. This is expected, as the 
audio-adaptive class token better incorporates complementary information from sound for the final activity classification, with a standard learnable class token the visual features will dominate the fusion inside the transformer.

\begin{table}[t!]
\centering
\resizebox{0.8\linewidth}{!}{
\begin{tabular}{lccc}
\toprule
 & \multicolumn{2}{c}{\textbf{Activities}} &\textbf{Overall} \\
 \cmidrule(lr){2-3} \cmidrule(lr){4-4}
\textbf{Model} & Silent & Audible & mAP (\%) $\uparrow$\\
\midrule
Visual encoder $\mathcal{V}(\cdot)$ & 23.2 & 22.7 & 23.1\\
Full model & 26.3 & 25.9 & 26.3\\
\bottomrule
\end{tabular}
}
\vspace{-0.8em}
\caption{\textbf{Benefit over silent and audible activities} on CharadesEgo. Our audio-adaptive model benefits both activity types.}
\vspace{-1em}
\label{tab:silent}
\end{table}


\begin{table*}[t!]
\centering
\resizebox{0.9\linewidth}{!}{
\begin{threeparttable}
\begin{tabular}{lccccccccccc}
\toprule
& \multicolumn{3}{c}{\textbf{Modality}} & \multicolumn{7}{c}{\textbf{EPIC-Kitchen Activity Recognition Across Domains}}\\
\cmidrule(lr){2-4} \cmidrule(lr){5-11} 
\textbf{Method} & RGB & Flow & Audio & D2 $\rightarrow$ D1 & D3 $\rightarrow$ D1 & D1 $\rightarrow$ D2 & D3 $\rightarrow$ D2 & D1 $\rightarrow$ D3 & D2 $\rightarrow$ D3 & \textit{Mean}\\
\midrule
\rowcolor{Gray}
\textbf{I3D backbone} & & & & &  &  &  &  &  &  \\
Source-only~\cite{munro2020multi} & $\checkmark$ & $\checkmark$ & & 42.5 & 44.3 & 42.0 & 56.3 & 41.2 & 46.5 & 45.5 \\
Munro and Damen \cite{munro2020multi}  & $\checkmark$&$\checkmark$& & 48.2 & 50.9 & 49.5 & 56.1 & 44.1 & 52.7 & 50.3 \\
Planamente \etal \cite{planamente2021cross}$^\dagger$ & $\checkmark$ & $\checkmark$ & $\checkmark$ & 48.5 & 50.9 & 49.7 & 56.3 & 44.8 & 52.5 & 50.5 \\
Yang \etal \cite{yang2021epic}$^\dagger$ & $\checkmark$ & $\checkmark$ & $\checkmark$ & 49.2 & 51.0 & 49.8 & 56.5 & 45.7 & 52.3 & 50.8 \\
Kim \etal \cite{iccv2021videoadaptation}  &$\checkmark$& $\checkmark$  & & 49.5 & 51.5 & 50.3 & 56.3 & 46.3 & 52.0 & 51.0 \\
Song \etal \cite{song2021spatio} &$\checkmark$ &$\checkmark$ & & 49.0 & \textbf{52.6} & 52.0 & 55.6 & 45.5 & 52.5 & 51.2 \\
\textit{\textbf{This paper} } & $\checkmark$ & $\checkmark$ & $\checkmark$ & \textbf{51.9} & 48.7 & \textbf{53.2} & \textbf{63.2} & \textbf{52.1} & \textbf{55.5} & \textbf{54.1}  \\
\rowcolor{Gray}
\textbf{SlowFast backbone} & & & &  &  &  &  &  &  &  \\
\textit{\textbf{This paper} } & $\checkmark$& $\checkmark$& $\checkmark$ & \textbf{59.3} & \textbf{59.1} & \textbf{59.5} & \textbf{69.1} & \textbf{54.8} & \textbf{64.3} & \textbf{61.0} \\
\bottomrule
\end{tabular}
\footnotesize{$^{\dagger}$~Based on our re-implementation using our features for RGB, flow and audio.}
\end{threeparttable}
}
\vspace{-0.5em}
\caption{\textbf{Activity recognition under scenery shift} on EPIC-Kitchens for the unsupervised domain adaptation setting. 
Our audio-adaptive model achieves state-of-the-art top-1 accuracy, and benefits from audio more than the audio-visual fusion methods used in prior works~\cite{planamente2021cross,yang2021epic}. 
Results increase further with a SlowFast backbone. More comparisons and modality-combinations are provided in the supplementary.}
\vspace{-0.8em}
\label{tab:epic_kitchens}
\end{table*}

\noindent \textbf{Benefit for silent activities.} In Table~\ref{tab:silent}, we demonstrate the effect of our 
full model on silent and audible activities separately. We focus on CharadesEgo since only 13 out of 157 classes have a characteristic sound (see supplementary). Our model obtains $\sim$3\% absolute increase for both silent and audible activities 
over a visual-only encoder. We conclude that audio is helpful for handling visual distribution shifts even for activities which do not have a characteristic sound.  

\noindent \textbf{Benefit for silent videos.} We have also tested our approach when the audio track is available for training but unavailable during inference. On EPIC-Kitchens, the audio-adaptive encoder achieves 50.7\% top-1 accuracy, still an improvement over visual encoder only (48.0\%). 
With both the audio-adaptive encoder and audio-infused recognizer, the result improves to 51.2\%. This indicates our approach effectively uses audio to help the visual encoder learn a more discriminative feature representation in the target domain, even when audio is absent during inference.

\begin{figure}[t!]
\centering 
\includegraphics[width=\linewidth,height=0.6\linewidth]{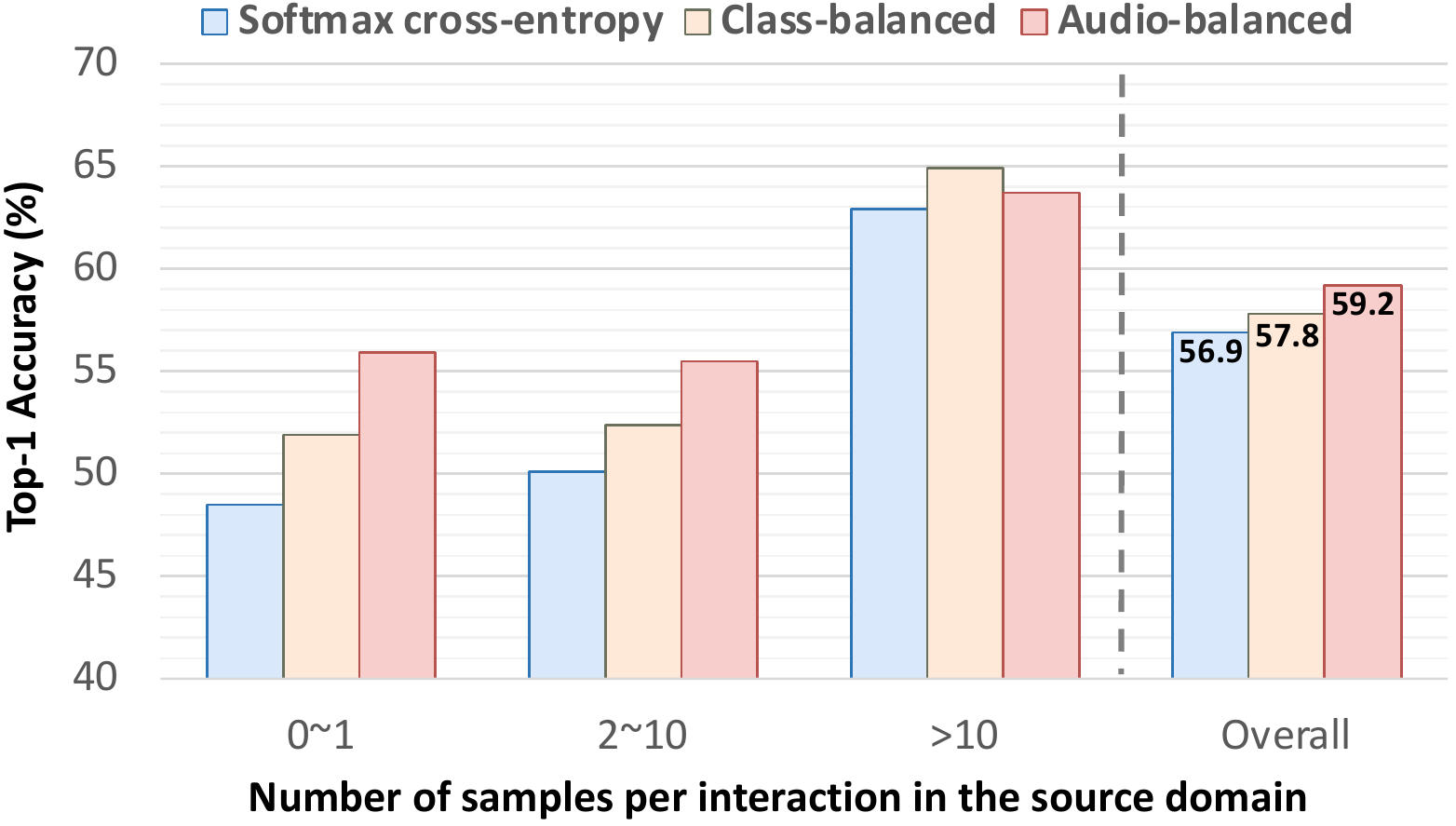}
\vspace{-1.5em}
\caption{\textbf{Benefit for the long-tail} on EPIC-Kitchens. Our audio-balanced loss learns rare activities in the source domain to generalize better to unknown activity distributions in the target domain. 
}
\label{fig:rare_interaction_figure}
\vspace{-1.1em}
\end{figure}

\noindent \textbf{Benefit for the long-tail.} 
In Figure~\ref{fig:rare_interaction_figure}, we demonstrate the benefit of audio-balanced learning towards activities that are rare in the source domain 
but are more frequent in the target domain. 
We use EPIC-Kitchens since it contains a long-tail of different object interactions (nouns) in each activity class (verb). We treat verb-noun pairs as frequent when they occur more than 10 times in the source domain, else they are considered rare. As the distribution of activities (verbs) changes across domains, the class-balanced loss~\cite{cui2019class} improves over the standard softmax cross-entropy loss. However, the domain shift also causes imbalance in the distribution of interactions (nouns). 
Because we balance the loss of each pseudo-interaction by clustering, our audio-balanced loss is especially helpful for the rare interactions (0-1 and 2-10 instances) where it obtains $\sim$3.5\% improvement. In comparison to the class-balanced loss we are slightly worse on frequent interactions, as we give higher weight to less common interactions. As interactions have a long-tail, our audio-balanced loss does result in an overall improvement.
%

\subsection{Comparison with State-of-the-Art}
\vspace{-0.3em}
\noindent \textbf{Scenery shift.} 
We first demonstrate the effectiveness of our approach for domain adaptation on EPIC-Kitchens, as defined by Munro and Damen~\cite{munro2020multi}. Here, different domains mean a change in scenery. The results are shown in Table~\ref{tab:epic_kitchens}. 
We first note that our approach gives $\sim$3\% improvement over the best performing prior works with the same I3D backbone. A further $\sim$7\% improvement can be gained from using SlowFast as the backbone~\cite{feichtenhofer2019slowfast}. There are several reasons for this improvement. First, our model utilizes the domain-invariant nature of audio signals to produce reliable pseudo-absent labels for the target domain video during training. This is particularly helpful for first-person videos where the activity may happen out of view. In addition, both RGB and Flow suffer from large appearance variance making it harder to guide domain-adaption through these modalities alone. Second, since the dataset has imbalanced label distributions, treating all the classes and interactions equally, as in prior works, results in inaccurate predictions when the semantic distribution shifts.

We also compare our full model with alternative audio-visual approaches proposed for cross-domain activity recognition~\cite{planamente2021cross, yang2021epic}. 
We let both of them use the same inputs, \ie, the 
features as outputted by the visual and audio encoders. 
Both of them use an adversarial loss to first align the visual features between domains and fuse visual and audio features or predictions afterwards. 
This causes the visual features to dominate the classification while the complementary information from sound may not be considered. 
Planamente \etal \cite{planamente2021cross} introduce an audio-visual loss, so the two modalities make a more balanced contribution towards the prediction. 
However, the audio predictions for silent activities are unreliable and harm their accuracy. 
Our model better combines the complementary information in the audio and visual modalities, effectively coping with many activities being silent.

\begin{figure}[t!]
\centering
\includegraphics[width=0.95\linewidth,height=0.7\linewidth]{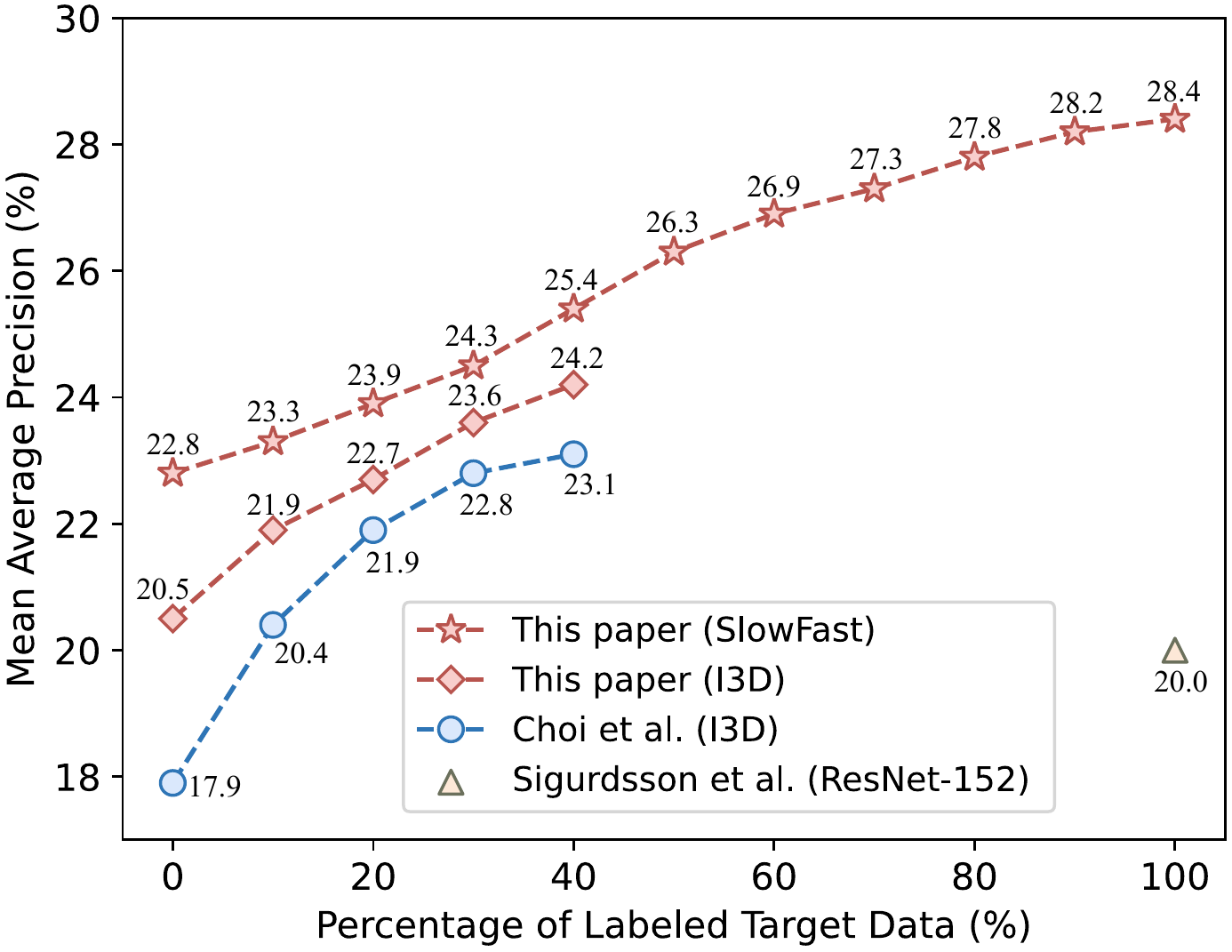}
\vspace{-0.5em}
\caption{\textbf{Activity recognition under viewpoint shift} on CharadesEgo. Using all source data, we compare with~\cite{choi2020unsupervised,sigurdsson2018actor} under varying amounts of labeled target training data. Our model obtains favourable results under all settings. 
}
\vspace{-1em}
\label{fig:CharadesEgo_comparison}
\end{figure}


\noindent \textbf{Viewpoint shift. }
In this comparison we consider viewpoint shift in CharadesEgo~\cite{sigurdsson2018actor}, following the semi-supervised setting of Choi \etal \cite{choi2020unsupervised}. Meaning we have some labeled target domain videos available during training. The results are shown in Figure~\ref{fig:CharadesEgo_comparison}. Our method achieves better results than Choi \etal \cite{choi2020unsupervised} with the same I3D RGB backbone~\cite{Kinetics}, for all amounts of labeled target videos. When adopting the SlowFast RGB backbone~\cite{feichtenhofer2019slowfast}, we again further improve performance for all settings. In the supplementary, we also provide a favorable  comparison with Li \etal \cite{li2021ego} under their fully-supervised setting. 
Since CharadesEgo contains paired first-person and third-person videos, we can test whether our method needs to see the same action instance from different viewpoints as in previous methods~\cite{sigurdsson2018actor} or whether it can make use of unpaired videos. When half of the paired videos from both views are used, we achieve a mAP of 29.9. When we use unpaired videos, the performance remains unchanged. We conclude our approach does not require paired training videos to be robust to viewpoint shift. 

\begin{figure}[t!]
\centering 
\includegraphics[width=1\linewidth,height=0.9\linewidth]{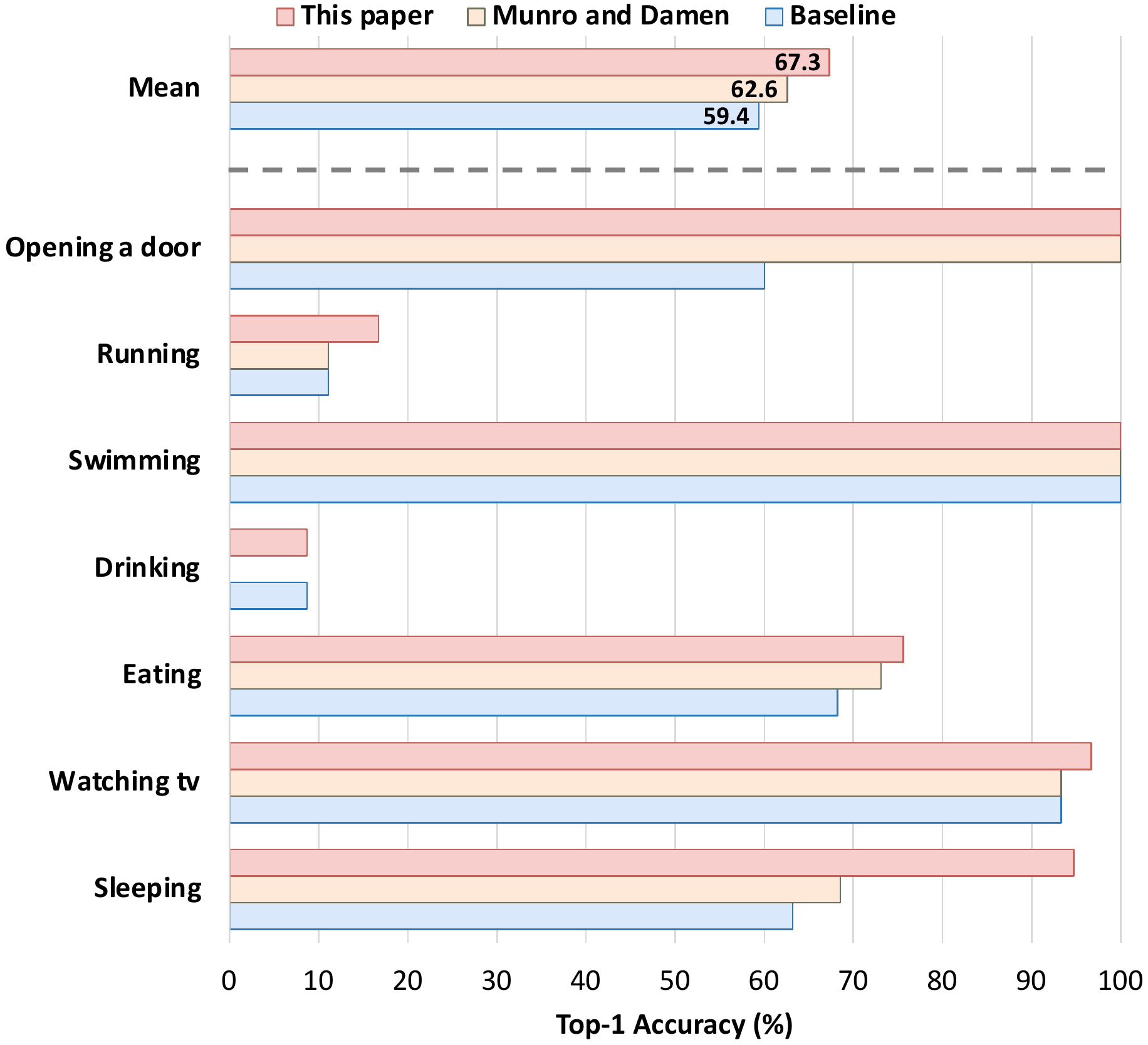}
\vspace{-1.8em}
\caption{\textbf{Activity recognition under actor shift} on our ActorShift dataset. When the visual similarities for the same activity are difficult to discover between domains, our model can use additional cues from sound 
to improve the recognition accuracy. 
}
\label{fig:actorshift}
\vspace{-1em}
\end{figure}

\noindent \textbf{Actor shift.} For this experiment, we use our ActorShift dataset and compare our model with the method by Munro and Damen~\cite{munro2020multi}, as their code is available. 
For fair comparison, we replace their I3D backbone with the same SlowFast backbone used for our model. 
We also show a baseline of the SlowFast model trained on source domain video only. 
The results are shown in Figure~\ref{fig:actorshift}. 
While the method proposed by Munro and Damen~\cite{munro2020multi} achieves good performance, our audio-adaptive approach better handles the large activity appearance variance caused by the shift in actors. For example, 
humans and animals sleep in visually different places and positions, while the sound of snoring or breathing is common to both. 
All models struggle with silent activities when there is both a large shift in appearance 
and a significant difference in sounds of activities between the domains, such as \emph{drinking} and \emph{running}.
We provide examples in the supplemental material, which are of interest for future work.

\section{Discussion}
\vspace{-0.3em}
\noindent\textbf{Limitations.} During training, our method needs videos from both source and target domains, and all should have an audio track with decent quality, limiting our approach to multi-modal video training sets. While audio at test-time is not required, it benefits activity recognition results considerably.

\noindent\textbf{Potential negative impact.}
When deployed our approach will have to record, store and process video and audio information related to human activities, which will have privacy implications for some application domains. 

\noindent\textbf{Conclusions.} 
We propose to recognize activities under domain shift with the aid of sound, using a novel audiovisual model. 
By leveraging the domain-invariant activity information within sound, our model improves over both silent and audible activities as well as rare activities in the source domain. 
Experiments on two domain adaptation benchmarks demonstrate that our approach has better adaptation ability than visual-only solutions and benefits from audio more than alternative audiovisual fusion methods used in prior works. 
We also show that our model better handles large activity appearance variance caused by the shift in actors. 

\noindent \textbf{Acknowledgement. }
This work is financially supported by the Inception Institute of Artificial Intelligence, the University of Amsterdam and the allowance Top consortia for Knowledge and Innovation (TKIs) from the Netherlands Ministry of Economic Affairs and Climate Policy.


{\small
\bibliographystyle{ieee_fullname}
\bibliography{egbib}
}
\clearpage

\appendix

\section{Implementation Details}
\label{appendix:impl}
\noindent \textbf{Visual encoder.}
On all datasets, we adopt the mmaction2~\cite{2020mmaction2} toolkit. 
The SlowFast~\cite{feichtenhofer2019slowfast} network is used to extract features from RGB frames for experiments on the EPIC-Kitchens, CharadesEgo and ActorShift datasets. 
We also extract features from the optical flow modality on EPIC-Kitchens by the slow-only network~\cite{feichtenhofer2019slowfast}. 
The networks are initialized with Kinetics pre-trained model weights. 
Such pre-training on a large dataset is common in domain adaptation, \eg, for images (ImageNet) \cite{ganin2015unsupervised,chen2018domain,tsai2019domain,tzeng2017adversarial} and videos (Sports-1M \cite{jamal2018deep}, Kinetics \cite{chen2020action,munro2020multi,choi2020shuffle,song2021spatio,iccv2021videoadaptation}).

\noindent \textbf{Audio encoder.} We adopt a ResNet-18~\cite{resnet} for all datasets and initialize the weights from the VGGSound pre-trained checkpoint~\cite{VGGSound}. 
The last residual block and final classification layer are further trained on each dataset before generating pseudo-absent labels and audiovisual fusion as detailed in Section~\textcolor{mydarkblue}{3}.

\noindent \textbf{Attention module. }
The attention module consists of eight
transformer encoder layers\cite{vaswani2017attention,devlin2018bert,vit} and the parameters are randomly initialized. The inputs are intermediate audio features from the audio encoder (conv3 of the ResNet-18 network). The output of the class token passes through one fully connected layer to obtain the attention vector for the visual encoder. 

\noindent \textbf{Audio-infused recognizer.} We use three transformer encoder layers with the same architecture as in~\cite{vit}. 
The parameters are also randomly initialized. 
The sequence dimension $D$ is set to 512 and each layer has 8 self-attention heads.

\noindent \textbf{Training objective.} On EPIC-Kitchens, We use a standard softmax cross-entropy loss as $\mathcal{L}$ in Eq.~\textcolor{mydarkblue}{2} and Eq.~\textcolor{mydarkblue}{8}. 
Since CharadesEgo aims for multi-label classification, the sigmoid cross-entropy loss is adopted as the $\mathcal{L}$ in Eq.~\textcolor{mydarkblue}{2} and Eq.~\textcolor{mydarkblue}{8}. 

\noindent \textbf{Inference.} When using a single modality, the output from the activity recognizer $\mathcal{R}(\cdot)$ is directly used as the final recognition prediction. 
On EPIC-Kitchens~\cite{munro2020multi}, when using both RGB and optical flow, we average the predictions from the two modalities as the final classification result, following prior works~\cite{munro2020multi,song2021spatio,iccv2021videoadaptation}.

\section{Audible Activities on CharadesEgo}
\label{appendix:audible_charadesego}
The 13 audible activities we select for the ablation in Table~\textcolor{mydarkblue}{3} are: 
\emph{someone is laughing}, \emph{someone is cooking something}, \emph{laughing at television}, \emph{closing a door}, \emph{talking on a phone/camera}, \emph{closing a window}, \emph{closing a refrigerator}, \emph{washing some clothes}, \emph{watching television}, \emph{washing their hands}, \emph{opening a window}, \emph{opening a door}, \emph{someone is sneezing}.

\begin{table}[t!]
\centering
\resizebox{1\linewidth}{!}{
\begin{tabular}{lcc}
\toprule
  & \multicolumn{1}{c}{\textbf{EPIC-Kitchens}} & \multicolumn{1}{c}{\textbf{CharadesEgo}} \\
\cmidrule(lr){2-2} \cmidrule(lr){3-3}
 \textbf{Labels for Target Domain} & Top-1 (\%) $\uparrow$ & mAP (\%) $\uparrow$  \\
\midrule
Visual-based hard pseudo labels & 51.7 & 23.9 \\
Visual-based pseudo-absent labels & 53.8 & 24.3 \\
Audio-based hard pseudo labels & 47.6 & 22.8 \\
Audio-based pseudo-absent labels & 55.7  & 25.0  \\ 
\bottomrule
\end{tabular}
}
\renewcommand\thetable{5}
\vspace{-1em}
\caption{\textbf{Ablation of pseudo-absent labels.} Using the pseudo-absent labels predicted by audio for absent-activity learning is more effective than the visual counterpart. Hard pseudo labels from either modality results in inferior performance. }
\label{tab:appendix_pseudo_absent}
\end{table}

\section{Ablation of Pseudo-Absent Labels}
\label{appendix:ablation_absent}
\noindent \textbf{Audio \textit{vs.} visual pseudo labels.} 
We introduce absent-activity learning in Section~\textcolor{mydarkblue}{3} to increase the discriminability of our audio-adaptive encoder in the target domain. 
The pseudo-absent labels are obtained from the audio encoder pre-trained on the source domain. 
We validate the effectiveness of this setting in Table~\ref{tab:appendix_pseudo_absent}. 
Here, we consider three alternatives. 
First, using a pre-trained visual encoder on the source domain to get ``Visual-based pseudo-absent labels" or ``Visual-based hard pseudo labels". In the latter a one-hot pseudo label for each video is obtained by taking the class with the highest probability. 
We can also create ``Audio-based hard pseudo labels" in the same way from the pre-trained audio encoder. 
Note that when using the hard pseudo labels, we adopt a standard classification loss, \ie softmax cross-entropy or sigmoid cross-entropy loss, for unlabeled target domain data, instead of the loss for absent-activity learning. 

Both visual-based pseudo-absent and hard pseudo labels result in inferior performance, since the visual activity appearance has larger variance across domains than audio and the visual encoder suffers from distribution shift. 
Since the audio predictions are unreliable for silent activities, using the hard pseudo labels from audio is worse than visual-based pseudo-labels. 
By contrast, our audio-based pseudo-absent labels provide reliable supervisory signals for the visual encoder adapting to the domain shift.

\noindent \textbf{Effect of $r$.} 
When we generate pseudo-absent labels for single-label classification on EPIC-Kitchens, $r$ classes with the lowest audio-based probabilities are treated as the absent activities to train our audio-adaptive encoder. In all other experiments $r{=}3$ is used. We illustrate the effect of $r$ in Table~\ref{tab:appendix_effect_of_r}. 
When $r$ equals 1, little supervision is provided for training the visual encoder in the target domain so that its adaptation ability degrades. 
With a large $r$, the pseudo-absent labels are noisy, since the audio predictions for silent activities are unreliable.
Overall, we consider $r{=}3$ to be the best trade-off. 

\begin{table}[t!]
\centering
\resizebox{0.5\linewidth}{!}{
\begin{tabular}{lcc}
\toprule
 \textbf{Value of $r$} & Top-1 (\%) $\uparrow$  \\
\midrule
1 & 52.3 \\
2 & 54.3  \\
3 & 55.7 \\
4 & 55.2 \\
5 & 54.1 \\
6 & 51.9 \\
\bottomrule
\end{tabular}
}
\renewcommand\thetable{6}
\vspace{-1em}
\caption{\textbf{Effect of $r$} for audio-based pseudo-absent labels on single-label classification with EPIC-Kitchens. While a small $r$ provides little supervision in the target domain, a large $r$ also degrades the performance due to the unreliable predictions for silent activities. 
$r{=}3$ results in the best performance. }
\label{tab:appendix_effect_of_r}
\end{table}

\begin{table}[t!]
\centering
\resizebox{0.5\linewidth}{!}{
\begin{tabular}{lcc}
\toprule
 \textbf{Value of $\gamma$} & mAP (\%) $\uparrow$  \\
\midrule
0.03 & 24.5 \\
0.04 & 24.8  \\
0.05 & 25.0 \\
0.06 & 24.7 \\
0.08 & 24.5 \\
0.1 &  24.1\\
\bottomrule
\end{tabular}
}
\renewcommand\thetable{7}
\vspace{-1em}
\caption{\textbf{Effect of $\gamma$} for audio-based pseudo-absent labels on multi-label classification with CharadesEgo. 
$\gamma{=}0.05$ results in the best performance. 
}
\label{tab:appendix_effect_of_gamma}
\end{table}

\noindent \textbf{Effect of $\gamma$.} 
For the multi-label classification, we assume the $(1-\alpha_k)\gamma$ percent videos with the lowest audio-based probabilities do not contain class $k$, where $\alpha_k$ is the percentage of videos containing class $k$ in the labeled source domain. 
Then, we obtain the pseudo-absent label for each video according to this rule. 
Here, we study the effect of $\gamma$ in Table~\ref{tab:appendix_effect_of_gamma}. Similar to the effect of $r$, $\gamma{=}0.05$ delivers the best result, while a smaller or larger $\gamma$ causes the performance to degrade slightly.

\noindent \textbf{Audio variance across domains.} The audio modality is more domain-invariant for distinguishing true negatives. When predicting absent activities in the source domain, both audio and visual classifiers achieve a high true negative rate on EPIC-Kitchens of 96.1\% and 97.2\%. 
In the target domain the audio remains robust with 95.6\% true negative rate, while that of the visual classifier degrades to 86.7\%.

\section{Ablation of Audio-Balanced Learning}
\setcounter{figure}{6}    

\label{appendix:ablation_balanced}
For our audio-balanced learning, we use audio features to cluster the samples inside each class in the source domain, and each cluster is treated as one type of interaction with objects or environments. 

\noindent \textbf{Effect of Elbow method.} 
The Elbow method~\cite{thorndike1953belongs} is adopted for determining the number of clusters for each class. It gives 5 to 12 clusters per class on both EPIC-Kitchens and CharadesEgo. 
Here, we compare its performance with a fixed number of clusters for all classes in Table~\ref{tab:appendix_elbow}. 
The Elbow method~\cite{thorndike1953belongs} results in the best performance. 
The reason is that some activity classes do not have large variance in the interactions, \eg, \emph{pouring}. 
For such an activity class, when using a large number of clusters, some clusters may contain samples with similar activity appearance to those in some other clusters of this class. 
Then, with audio-balanced learning, the visual encoder may pay more attention to these `redundant' clusters during training, as they contain less samples, and over-fit to the samples in these clusters. 
Besides, if all the classes adopt a small number of clusters, some rare interactions with objects or environments in the source domain cannot be well learned and thus the overall accuracy of the model degrades on the target domain. 

\begin{table}[t!]
\centering
\resizebox{0.9\linewidth}{!}{
\begin{tabular}{lcc}
\toprule
  & \multicolumn{1}{c}{\textbf{EPIC-Kitchens}} & \multicolumn{1}{c}{\textbf{CharadesEgo}} \\
\cmidrule(lr){2-2} \cmidrule(lr){3-3}
 \textbf{Number of Clusters} & Top-1 (\%) $\uparrow$ & mAP (\%) $\uparrow$  \\
\midrule
4 & 52.9 & 23.5 \\
5 & 53.4 & 23.9 \\
6 & 53.8 & 24.2 \\
7 & 54.3 & 24.5 \\
8 & 53.9 & 24.1 \\
9 & 53.6 & 23.8 \\
10 & 52.7 & 23.3 \\
Elbow method~\cite{thorndike1953belongs} & 55.7  & 25.0  \\ 
\bottomrule
\end{tabular}
}
\renewcommand\thetable{8}
\vspace{-1em}
\caption{\textbf{Effect of Elbow method} on our audio-adaptive encoder. Using a fixed number of clusters results in inferior performance compared to the Elbow method. }
\label{tab:appendix_elbow}
\end{table}

\begin{table}[t!]
\centering
\resizebox{0.7\linewidth}{!}{
\begin{tabular}{lcc}
\toprule
  & \multicolumn{1}{c}{\textbf{EPIC-Kitchens}} & \multicolumn{1}{c}{\textbf{CharadesEgo}} \\
\cmidrule(lr){2-2} \cmidrule(lr){3-3}
 \textbf{Modality} & Top-1 (\%) $\uparrow$ & mAP (\%) $\uparrow$  \\
\midrule
Visual & 53.4 & 24.1 \\
Audio & 55.7  & 25.0  \\ 
\bottomrule
\end{tabular}
}
\renewcommand\thetable{9}
\vspace{-1em}
\caption{\textbf{Audio \textit{vs.} visual features for clustering} in the audio-balanced learning for our audio-adaptive encoder. Using audio features is preferred over the visual features and delivers 2.3\% top-1 accuracy and 0.9\% mAP advantage. }
\label{tab:appendix_clustering}
\end{table}

\begin{figure}[t!]
\centering
\includegraphics[width=\linewidth,height=0.6\linewidth]{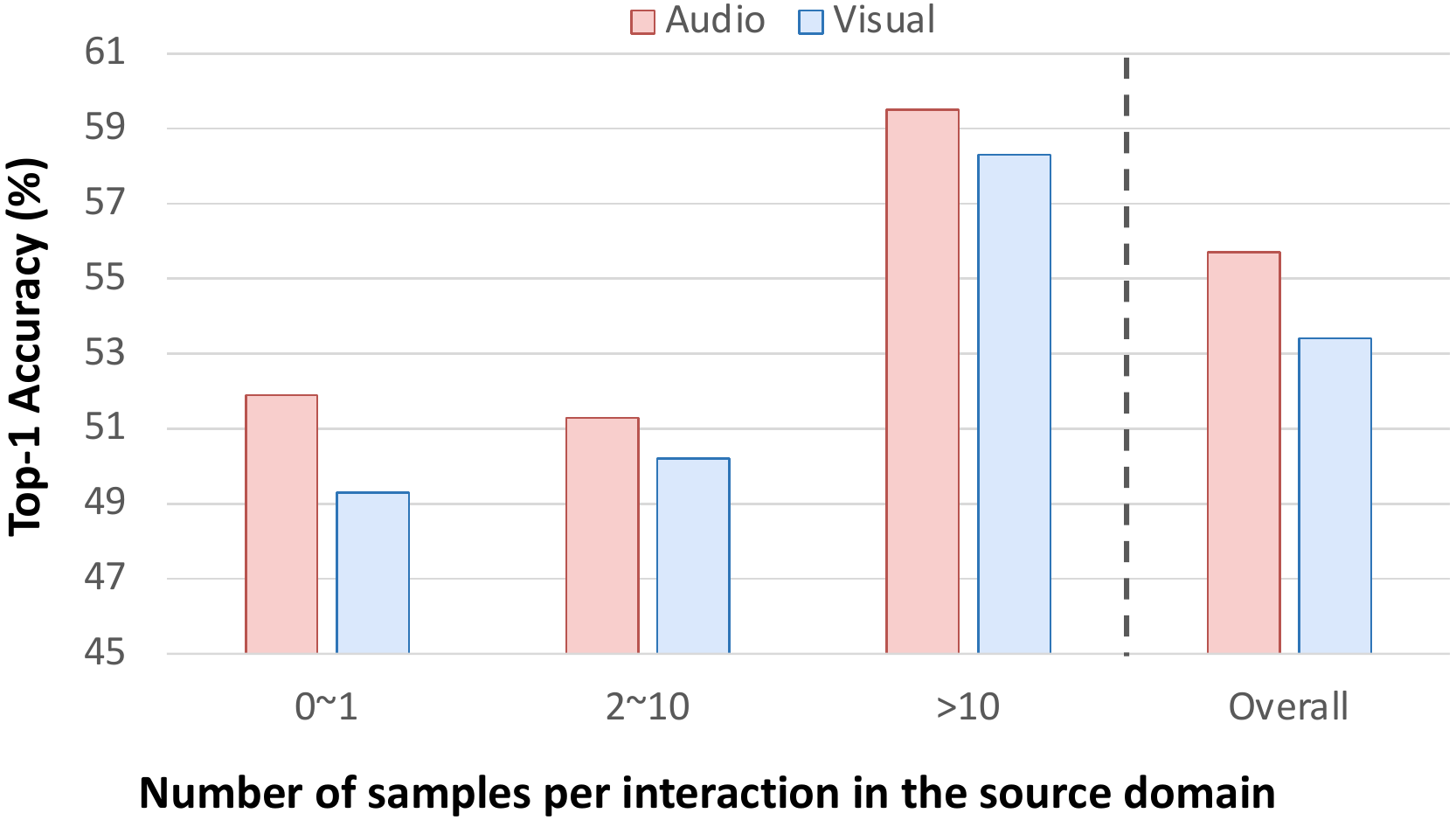}
\vspace{-1.5em}
\caption{\textbf{Audio \textit{vs.} visual features for long-tail} on EPIC-Kitchens. Clustering by audio features to identify rare and frequent activities for balanced learning is preferred over visual features under semantic distribution shift. 
}
\label{fig:appendix_rare_interaction_figure}
\end{figure}

\noindent \textbf{Audio \textit{vs.} visual features for clustering.} 
As an alternative to our audio-based clustering in Section~\textcolor{mydarkblue}{3}, we could instead rely on visual features for clustering. 
However, since visual features are sensitive to appearance changes, such as the background color, we observe some of the resulting clusters may mainly contain videos with similar backgrounds, rather than a specific type of interaction. 
We compare the performance per modality in Table~\ref{tab:appendix_clustering}. 
Clustering by audio outperforms the visual counterpart by 2.3\% top-1 accuracy on EPIC-Kitchens and 0.9\% mAP on CharadesEgo. 

We also show the difference in performance towards rare activities in EPIC-Kitchens when using visual-based clustering 
in Figure~\ref{fig:appendix_rare_interaction_figure}. 
Using audio for clustering delivers higher accuracy towards rare activities than the visual counterpart. 
We also observe that clustering by the audio modality is better than clustering by visual features on frequent activities. 
This is because the model with visual clustering may focus more on rare differences in backgrounds that have no effect on the activity class. 
We conclude that forcing the model to learn in a balanced way towards different types of interactions clustered by the audio can better handle semantic distribution shift than the visual counterpart.

\noindent \textbf{Visualizations of clusters.} 
Some examples from the clusters are provided in the supplementary video. We observe that each cluster by the audio features tends to contain similar objects or environments the actors interact with. For example, for the \emph{putting} activity in EPIC-Kitchens, one cluster has a bias towards food, one consists mainly of plates and another one prefers kitchen utensils. 
Similar phenomena also exist in  CharadesEgo. For the \emph{eating} activity, the actors are commonly watching television in the living room when eating in one cluster. 
Kitchens or bedrooms with silent environments are frequent sites for eating in another cluster. 
There is also one cluster in which there are several people chatting in each video while the actor of interest is eating. 
We conclude that when finding rare activities by clustering, using audio features is reliable.

\section{Ablation of Audio-Infused Recognizer}
\label{appendix:ablation_class_token}
\noindent \textbf{Audio-adaptive class token.} In Section~\textcolor{mydarkblue}{3}, we obtain the audio-adaptive class token from a series of activity sound feature vectors, considering both audio and visual features. 
Alternatively, we can directly conduct global average pooling on the audio features from the audio encoder and treat the resulting feature vector as the class token. 
Then, the audio-infused activity recognizer can be jointly trained with the last residual block of the audio encoder. 
The performance comparison is shown in Table~\ref{tab:appendix_class_token}. 
Since joint training includes more parameters than training the recognizer only, the model suffers from over-fitting~\cite{wang2020makes} and results in 1.4\% top-1 accuracy and 0.7\% mAP drops, compared to using activity sound feature vectors. 
However, if we fix the audio encoder and only train the recognizer, the performance degrades dramatically. 
This is expected as audio and visual features come from different feature distributions and cannot ensure effective audio-visual interaction by the attention inside the recognizer. 
We conclude our audio-adaptive class token from the activity sound feature vectors is superior compared to the original audio features from the audio encoder.

\begin{table}[t!]
\centering
\resizebox{1\linewidth}{!}{
\begin{threeparttable}
\begin{tabular}{lcc}
\toprule
  & \multicolumn{1}{c}{\textbf{EPIC-Kitchens}} & \multicolumn{1}{c}{\textbf{CharadesEgo}} \\
\cmidrule(lr){2-2} \cmidrule(lr){3-3}
 \textbf{Source of class token} & Top-1 (\%) $\uparrow$ & mAP (\%) $\uparrow$  \\
\midrule
Audio classification prediction & 57.5 & 25.6 \\
Audio features & 57.8 & 25.6 \\
Audio features$^{\dagger}$ & 55.9 & 25.2 \\
Activity sound feature vectors & 59.2  & 26.3  \\ 
\bottomrule
\end{tabular}
\footnotesize{$^{\dagger}$~Only training the audio-infused activity recognizer.}\\
\end{threeparttable}
}
\renewcommand\thetable{10}
\vspace{-1em}
\caption{\textbf{Audio-adaptive class token ablation. }Obtaining the audio-adaptive class token from activity sound feature vectors is preferred over using the original audio feature vector or classification prediction from the audio encoder as the class token. }
\label{tab:appendix_class_token}
\end{table}

\noindent \textbf{Recognizer \textit{vs.} simple classifier}. 
To further justify our audio-infused recognizer, we test an alternative classifier that uses the audio-attention weights $h$ concatenated with the visual feature as input to a fully connected layer. It scores 55.9\% top-1 accuracy on EPIC-Kitchens and 24.8\% mAP on CharadesEgo. Worse than our 59.2\% and 26.3\%.

\begin{table*}[t!]
\centering
\resizebox{0.99\linewidth}{!}{
\begin{tabular}{lccccccccccc}
\toprule
& \multicolumn{3}{c}{\textbf{Modality}} & \multicolumn{7}{c}{\textbf{EPIC-Kitchen Activity Recognition Across Domains}}\\
\cmidrule(lr){2-4} \cmidrule(lr){5-11} 
\textbf{Method} & RGB & Flow & Audio & D2 $\rightarrow$ D1 & D3 $\rightarrow$ D1 & D1 $\rightarrow$ D2 & D3 $\rightarrow$ D2 & D1 $\rightarrow$ D3 & D2 $\rightarrow$ D3 & \textit{Mean}\\
\midrule
\textit{\textbf{This paper} } & & & $\checkmark$& 37.2 & 38.4 & 40.7 & 44.7 & 46.0 & 47.0 & 42.3 \\
\rowcolor{Gray}
\textbf{I3D Architecture} & & & & &  &  &  &  &  &  \\
Munro and Damen \cite{munro2020multi}  & $\checkmark$&$\checkmark$& & 48.2 & 50.9 & 49.5 & 56.1 & 44.1 & 52.7 & 50.3 \\
Kim \etal \cite{iccv2021videoadaptation}  &$\checkmark$& $\checkmark$  & & 49.5 & 51.5 & 50.3 & 56.3 & 46.3 & 52.0 & 51.0 \\
Song \etal \cite{song2021spatio} &$\checkmark$ &$\checkmark$ & & 49.0 & \textbf{52.6} & 52.0 & 55.6 & 45.5 & 52.5 & 51.2 \\
\textit{\textbf{This paper} } & & $\checkmark$ & & 44.5 & 44.7 & 50.8 & 55.3 & 42.1 & 50.2 & 47.9 \\
\textit{\textbf{This paper} } & & $\checkmark$ & $\checkmark$ & 48.7 & 48.3 & 52.3 & 60.9 & 49.2 & 53.1 & 52.1 \\ 
\textit{\textbf{This paper} } & $\checkmark$& & & 38.1 & 37.8 & 39.2 & 47.9 & 42.1 & 41.8 & 41.1 \\
\textit{\textbf{This paper} } & $\checkmark$ & $\checkmark$& & 43.7 & 42.6 & 45.7 & 50.2 & 43.8 & 52.3 & 46.4 \\
\textit{\textbf{This paper} } & $\checkmark$ & & $\checkmark$ & 49.5 & 43.5 & 49.2 & 54.6 & 46.6 & 50.0 & 48.9 \\
\textit{\textbf{This paper} } & $\checkmark$ & $\checkmark$ & $\checkmark$ & \textbf{51.9} & 48.7 & \textbf{53.2} & \textbf{63.2} & \textbf{52.1} & \textbf{55.5} & \textbf{54.1} \\
\rowcolor{Gray}
\textbf{SlowFast Architecture} & & & &  &  &  &  &  &  &  \\
\textit{\textbf{This paper} } & & $\checkmark$& & 48.5 & 49.2 & 50.4 & 54.1 & 44.1 & 52.8 & 49.8 \\
\textit{\textbf{This paper} } & $\checkmark$ & & & 52.4 & 51.1 & 51.8 & 55.9 & 45.7 & 53.6 & 51.8 \\
\textit{\textbf{This paper} } & &$\checkmark$ &$\checkmark$ & 50.8 & 51.5 & 52.7 & 60.7 & 48.4 & 57.7 & 53.6 \\
\textit{\textbf{This paper} } & $\checkmark$ & $\checkmark$& & 53.9 & 54.3 & 53.5 & 58.7 & 47.9 & 55.2 & 53.9 \\
\textit{\textbf{This paper} } &$\checkmark$ & &$\checkmark$ & 57.9 & 55.9 & 58.4 & 64.3 & 54.2 & \textbf{64.3} & 59.2 \\ 
\textit{\textbf{This paper} } & $\checkmark$& $\checkmark$& $\checkmark$ & \textbf{59.3} & \textbf{59.1} & \textbf{59.5} & \textbf{69.1} & \textbf{54.8} & \textbf{64.3} & \textbf{61.0} \\
\bottomrule
\end{tabular}
}
\renewcommand\thetable{13}
\vspace{-0.8em}
\caption{\textbf{Modality ablation under scenery shift} on EPIC-Kitchens for the unsupervised domain adaptation setting. Relying on either audio or visual modality results in inferior performance, while our audio-adaptive models achieve state-of-the-art accuracy. }
\label{tab:appendix_epic_kitchens}
\end{table*}

\begin{table}[t!]
\centering
\resizebox{0.4\linewidth}{!}{
\begin{tabular}{lc}
\toprule
 \textbf{Depth} & Top-1 (\%) $\uparrow$  \\
\midrule
5 & 55.0 \\
6 & 55.3  \\
7 & 55.3 \\
8 & 55.7 \\
9 & 55.5 \\
10 & 55.3 \\
11 & 55.2 \\
12 & 54.9 \\
\bottomrule
\end{tabular}
}
\renewcommand\thetable{11}
\vspace{-0.8em}
\caption{\textbf{Effect of depth} for the audio-adaptive encoder on EPIC-Kitchens. The performance is not very sensitive to the depth and using 8 layers results in the best performance. }
\label{tab:appendix_effect_of_depth_att}
\end{table}

\begin{table}[t!]
\vspace{-1em}
\centering
\resizebox{0.4\linewidth}{!}{
\begin{tabular}{lc}
\toprule
 \textbf{Depth} & Top-1 (\%) $\uparrow$  \\
\midrule
1 & 56.8 \\
2 & 57.3  \\
3 & 59.2 \\
4 & 59.0 \\
5 & 58.4 \\
6 & 58.1 \\
\bottomrule
\end{tabular}
}
\renewcommand\thetable{12}
\vspace{-0.8em}
\caption{\textbf{Effect of depth} for the audio-infused recognizer on EPIC-Kitchens. 
Increasing depth to 3 layers is effective, then performance plateaus.}
\vspace{-0.8em}
\label{tab:appendix_effect_of_depth_recog}
\end{table}


\section{Effect of Depth in Transformer Modules}
\label{appendix:depth}

\noindent \textbf{Attention module. }
Our attention module consists of eight transformer encoder layers. 
With more layers, the module may suffer from over-fitting. By contrast, only using a few layers will lead to under-fitting. We study the effect of depth on our audio-adaptive encoder by setting it in the range of $[5,12]$, and the results on EPIC-Kitchens with RGB and audio modalities are shown in Table~\ref{tab:appendix_effect_of_depth_att}. 
Performance is not very sensitive to the depth, using 8 layers represents the best empirical trade-off.

\noindent \textbf{Audio-infused recognizer.} Our audio-infused recognizer contains 3 transformer layers. 
Similar to the attention module, more layers lead to over-fitting while less layers result in under-fitting. 
We study the effect of depth on  EPIC-Kitchens with RGB and audio modalities and the results are shown in Table~\ref{tab:appendix_effect_of_depth_recog}. 
Increasing the depth to 3 layers is effective, then performance plateaus.

\section{Modality Ablation on EPIC-Kitchens}
\label{appendix:epic_kitchens}
We provide more modality-combinations in Table~\ref{tab:appendix_epic_kitchens}. 
The audio encoder alone can achieve only 42.3\% top-1 accuracy, since it cannot predict accurately on silent activities. 
We also consider using the visual modalities only, \ie RGB and optical flow, and modify our approach correspondingly. 
To be specific, the pseudo-absent labels are determined by the pre-trained visual encoder in the source domain. 
Visual features are used for clustering in the audio-balanced learning as well as generating the attention for themselves. 
The activity recognizer also takes visual features only as inputs along with a learnable class token as in \cite{vit}. 
The visual-only versions of our approach achieve inferior performance since the activity appearance suffers from large variances under domain shift. 
However, our audio-adaptive model with either RGB or optical flow modality delivers competitive accuracy with the aid of the domain-invariant information within sound. 

\section{Additional Comparison on CharadesEgo}
\label{appendix:charadesego}
Li~\etal recently proposed a supervised-only approach~\cite{li2021ego} for model pre-training to be better suited for downstream tasks with egocentric videos. We find our approach profits from their features as well. On its own the approach from Li~\etal achieves 30.6 mAP on CharadesEgo. Combined with our audio-based attention and audio-infused recognizer, we improve this to 31.9 mAP.

\section{Fusion Benefit}
\label{appendix:fusion}
\begin{table}[t!]
\centering
\resizebox{0.6\linewidth}{!}{
\begin{tabular}{lc}
\toprule
 \textbf{Method} & Top-1 (\%) $\uparrow$ \\
\midrule
\rowcolor{Gray}
\textbf{Sight or Sound} & \\
Visual-only & 48.0 \\
Audio-only & 42.3 \\
\rowcolor{Gray}
\textbf{Within-domain fusion} & \\
Late fusion &  47.2 \\
Lee \etal~\cite{lee2021crossattentional} & 50.8 \\  
Tian \etal~\cite{tian2020unified} & 50.0 \\
Nagrani \etal~\cite{nagrani2021attention} & 51.1 \\
Gabeur \etal\cite{gabeur2020multi} & 51.3\\
\rowcolor{Gray}
\textbf{Cross-domain fusion} & \\
\textbf{\textit{This paper}} & \textbf{59.2} \\
\bottomrule
\end{tabular}
}
\vspace{-0.8em}
\renewcommand\thetable{14}
\caption{\textbf{Fusion benefit.} Experiments performed on EPIC-Kitchens with the same RGB and audio modalities. Although utilizing within-domain fusion methods can achieve good performance, our approach provides more effective cross-modal interaction under domain shift. 
}
\vspace{-0.5em}
\label{tab:comparison_audiovisual_fusion}
\end{table}

We also compare our full audio-adaptive model with alternative audio-visual fusion approaches originally intended for within-domain activity recognition~\cite{lee2021crossattentional, tian2020unified, gabeur2020multi, nagrani2021attention} on EPIC-Kitchens. 
We use either publicly available implementations or re-implement ourselves and let them all use the same inputs, \ie, the 
features as outputted by the visual and audio encoders. The results are shown in Table~\ref{tab:comparison_audiovisual_fusion}. 
We denote simple averaging of the classification predictions from both encoders by ``Late fusion". 
As most activities are silent, the audio predictions are unreliable and degrade the performance when combined with visual predictions via late fusion. 
Although the cross-modal interaction methods proposed in~\cite{lee2021crossattentional,tian2020unified,gabeur2020multi,nagrani2021attention} are designed for within-domain activity recognition, they still achieve good performance compared to a visual-only encoder with 48.0\% top-1 accuracy. 
This is expected as several samples in the target domain may not contain a large domain shift, so the audio-visual correspondences from the source domain will also be encountered in the target domain. 
Our full audio-adaptive model allows for an even more effective cross-modal interaction under domain shift.
This is because our audio-adaptive encoder and audio-infused recognizer alleviate the dependence on searching cross-modal correspondences for classification and instead rely on the domain-invariant activity information within sound to obtain a more discriminative visual feature representation in the target domain.

\begin{figure}[t!]
    \centering
    \includegraphics[width=0.95\linewidth,height=0.75\linewidth]{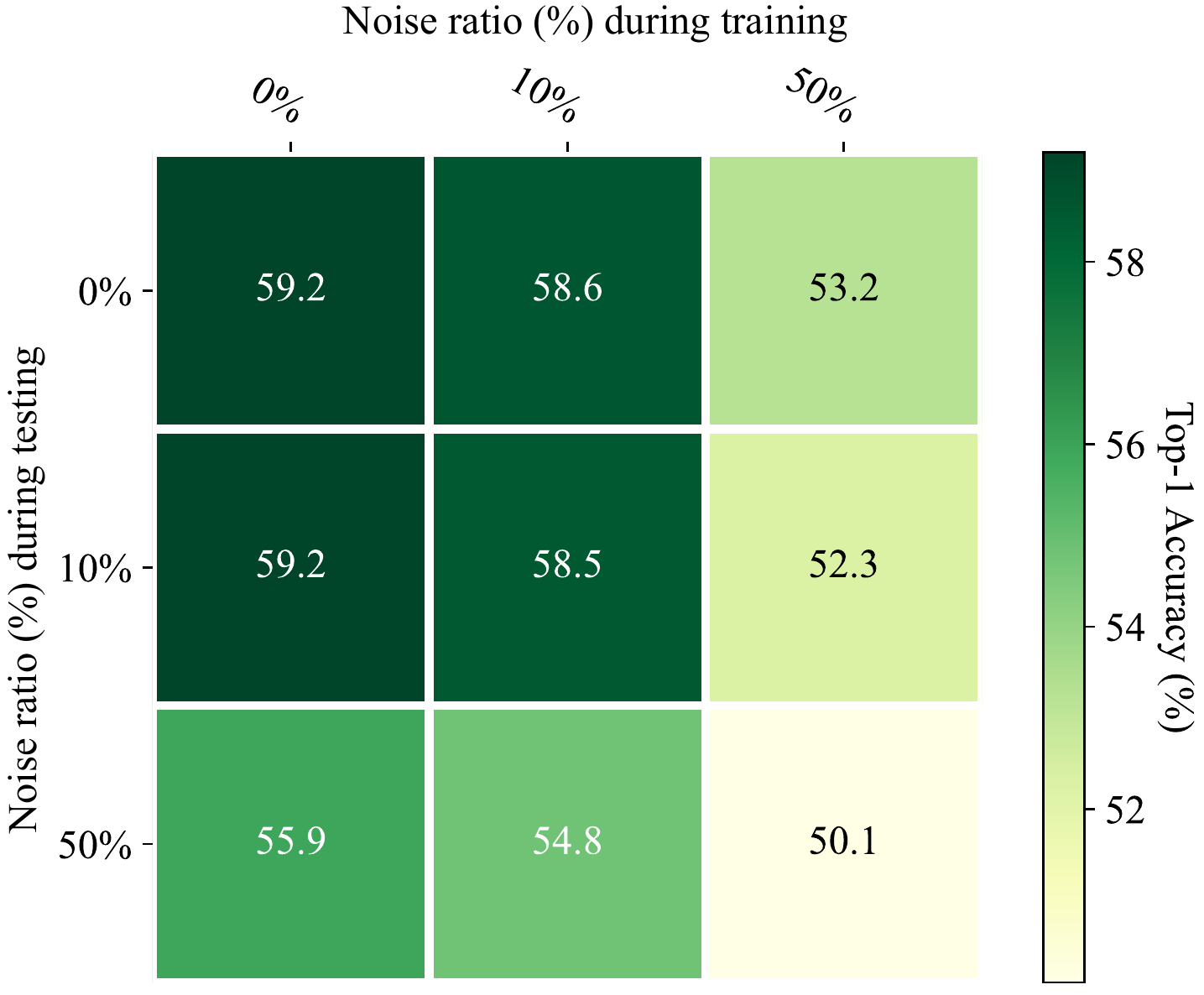}
    \caption{\textbf{Impact of audio quality} on EPIC-Kitchens. Our model remains robust up to 10\% irrelevant sound during training and/or testing. With more noise, especially during training, performance starts to suffer. }
    \label{fig:appendix_audio_quality}
\end{figure}

\section{Audio Quality Assumption}

Throughout our work, we assume the audio track accompanying a video is of decent quality. To measure the impact of audio quality, we mix the audio track of each video with the audio from another randomly sampled video. In Figure~\ref{fig:appendix_audio_quality} we vary the noise ratio in train and test and measure the top-1 accuracy (\%) on EPIC-Kitchens. Our model remains robust up to 10\% irrelevant sound during training and/or testing. With more noise, especially during training, performance starts to suffer.

\begin{figure}[t!]
    \centering
    \includegraphics[width=1\linewidth]{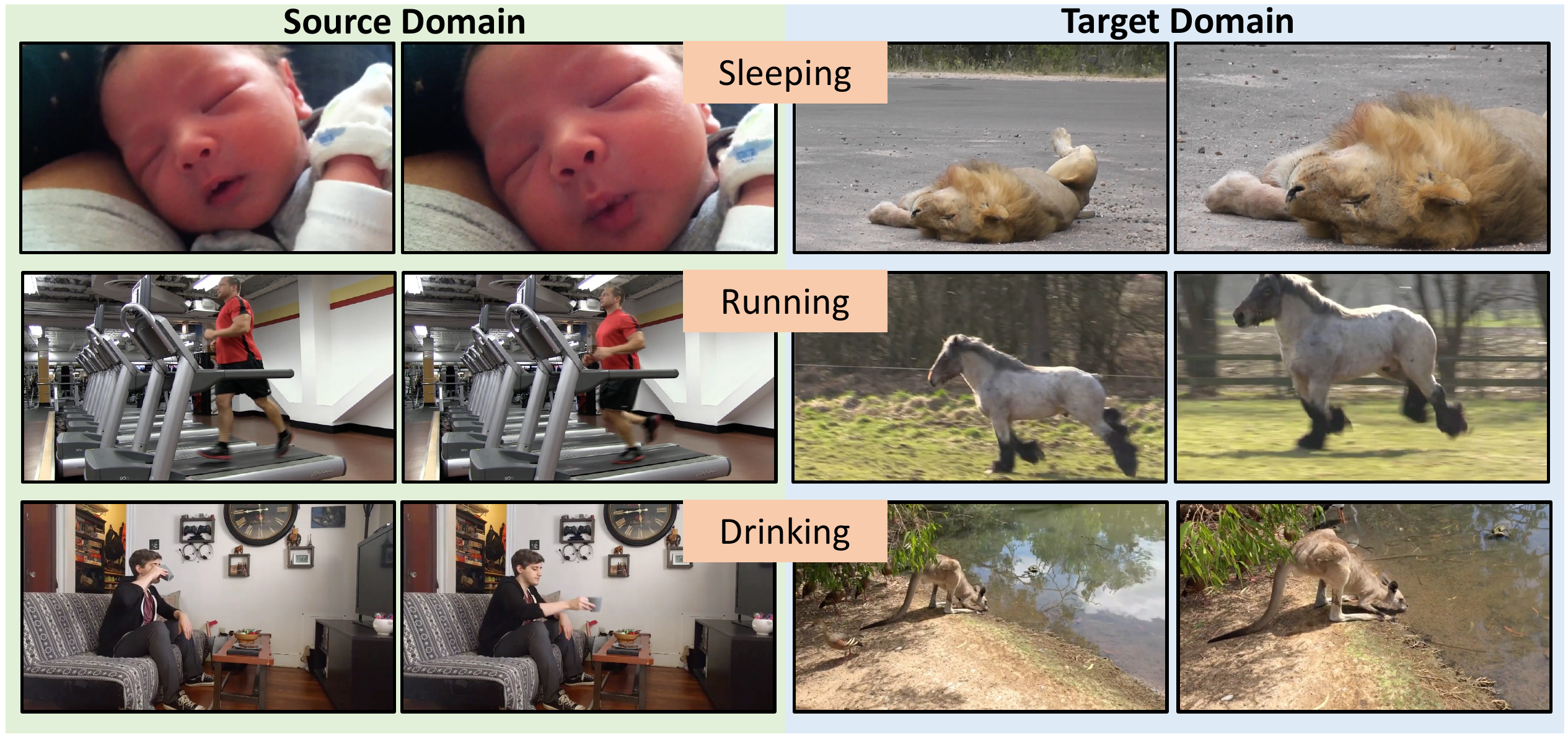}
    \caption{\textbf{Example activities from the ActorShift dataset}.}
    \label{fig:appendix_actorshift}
\end{figure}

\section{Details of ActorShift Dataset}
When constructing the dataset, we first select 7 Kinetics classes for which there are dozens of videos of animals performing these actions on YouTube. The videos are collected by querying YouTube with `animals' prepended to the verb of the action class, ~\textit{e.g.,} ``animals sleeping''. 
We discard videos with music and only keep those with the original animal sounds. Videos with the animal out-of-view are also rejected. 
The final dataset covers a wide range of animal species including dog, deer, koala, cat, alpaca, lion, tiger, kangaroo, loris, raccoon, rabbit, elephant, monkey, panda, horse, duck, bird, snail, cow, chinchilla, marmot, lizard, hedgehog, bat, tortoise, squirrel, giraffe, goose and fox. 
Some examples are shown in Figure~\ref{fig:appendix_actorshift}. 

\section{Supplementary Video}
\label{appendix:visualizations}
We provide more visualizations in the supplementary video on our project page: \url{https://xiaobai1217.github.io/DomainAdaptation}. 
It includes examples about the pseudo-absent labels for absent-activity learning and the clusters generated by audio features for audio-balanced learning. 
We also compare the predictions between a visual-only encoder and our audio-adaptive encoder, as well as the benefit brought by our audio-infused recognizer. 
Some failure cases on the ActorShift dataset are also shown, where the domain shift exists in both the visual and audio modalities. 

\end{document}